\documentclass[letterpaper]{article} 
\usepackage{aaai2027}  
\usepackage{adjustbox}
\usepackage{graphicx}
\usepackage{booktabs}
\usepackage{algorithm}
\usepackage{algorithmic}
\usepackage{adjustbox}
\usepackage{booktabs}
\usepackage{bibentry}
\usepackage{multirow}
\usepackage{graphicx}

\usepackage{amsmath}
\usepackage{changepage}
\usepackage{amssymb}
\usepackage[table]{xcolor} 
\usepackage{booktabs}      
\usepackage{multirow}      
\usepackage{bm}
\usepackage[hyphens]{url}  
\usepackage{graphicx} 
\def\UrlFont{\rm}  
\usepackage{natbib}  
\usepackage{caption} 
\usepackage{algorithm}
\usepackage{algorithmic}

\usepackage{newfloat}
\usepackage{listings}
\DeclareCaptionStyle{ruled}{labelfont=normalfont,labelsep=colon,strut=off} 
\floatstyle{ruled}
\newfloat{listing}{tb}{lst}{}
\floatname{listing}{Listing}

\usepackage{booktabs}

\title{CORMA: Constraint-Routed Decision Mamba for Generalizable \\ Multi-Constraint Offline Safe Reinforcement Learning}
\author{
Changxu Ni\textsuperscript{\rm 1},
Feiran You\textsuperscript{\rm 3},
Yongan Guo\textsuperscript{\rm 2},\\
Hongyang Du\textsuperscript{\rm 3},
Kang Wei\textsuperscript{\rm 1},
Le Liang\textsuperscript{\rm 1},
Yiyang Ni\textsuperscript{\rm 2},
Jun Li\textsuperscript{\rm 1}
}

\affiliations{
\textsuperscript{\rm 1}Southeast University\\
\textsuperscript{\rm 2}Nanjing University of Posts and Telecommunications\\
\textsuperscript{\rm 3}The University of Hong Kong
}

\begin{document}

\maketitle

\begin{abstract}
Offline safe reinforcement learning (RL)  learns safety-aware policies from static datasets while satisfying safety constraints during deployment. Motivated by the strong trajectory-modeling capabilities of Transformers, recent Decision Transformer (DT)-based methods formulate offline safe RL as conditional sequence modeling by conditioning action generation on return-to-go (RTG) and cost-to-go (CTG) tokens, thereby enabling adaptation to varying safety budgets. However, existing DT-based methods are developed for single-constraint tasks. For tasks with multiple constraints, generalizing learned policies to unseen combinations of safety budgets remains challenging due to the conflicting requirements imposed by heterogeneous constraints. Moreover, the quadratic complexity of Transformer architectures significantly increases the computational overhead in multi-constraint tasks, as representing additional constraints requires more CTG tokens.
Recently, Mamba has emerged as an efficient sequence-modeling backbone, offering linear computational complexity and strong generalization capabilities. In this paper, we propose CORMA, a constraint-routed decision Mamba framework for generalizable multi-constraint offline safe RL. CORMA decomposes policy inference into constraint-aware RTG estimation and adaptive action generation, where constraint-specific experts model feasible RTG distributions and a router-guided product-of-experts mechanism fuses these distributions to provide safety-aware guidance for a Mamba-based action decoder. A three-stage optimization strategy is designed to improve training stability. Extensive experiments on OSRL benchmarks demonstrate that CORMA achieves superior reward-safety trade-offs, reduces inference latency by 37$\%$ compared with DT-based methods, and exhibits stronger zero-shot generalization to unseen safety-budget combinations than offline safe RL baselines.
\end{abstract}

\section{Introduction}
Reinforcement learning (RL) has achieved remarkable success in complex decision-making tasks (Hansen et al., 2024; Wang et al., 2024). However, in safety-critical domains such as autonomous driving and industrial robot control, agents must maximize task rewards while satisfying safety constraints (Wei et al., 2024; Wachi et al., 2024). Safe RL (Gao et al., 2024; Hazra et al., 2025) addresses this problem through constrained Markov decision processes (CMDPs), where policies are optimized subject to predefined safety budgets (Liu et al., 2023a; Zhao et al., 2023). Most existing methods, however, require online interactions that may expose the agent or environment to unacceptable risks. Offline safe RL avoids such unsafe exploration by learning entirely from static datasets without additional environmental interactions (Alles et al., 2024; Gong et al., 2025).
\begin{figure}[t]
    \centering
    \includegraphics[width=0.47    \textwidth]{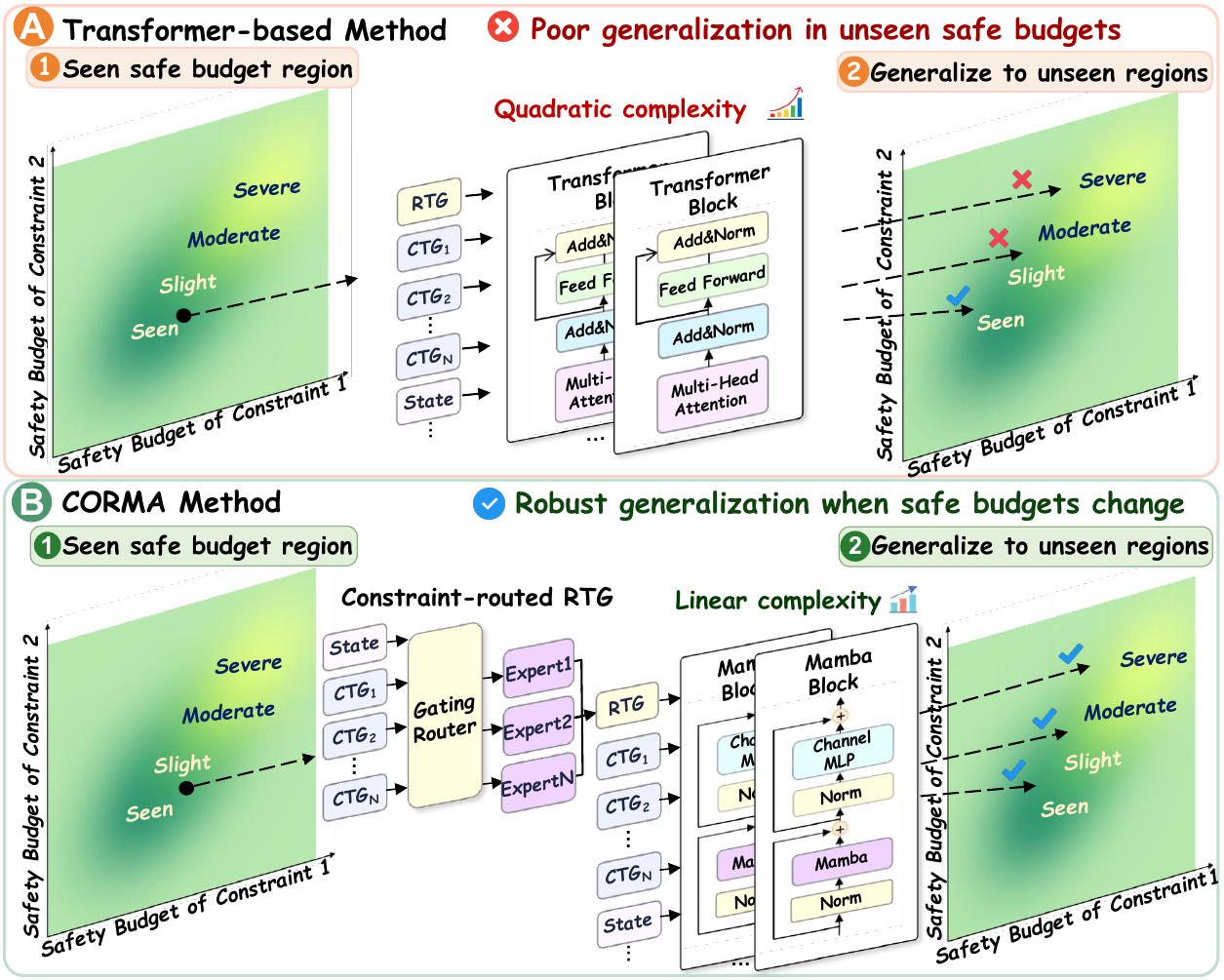} 
    \caption{Illustration of the challenges. After pre-training on seen safety-budget regions, Transformer-based sequence models struggle under unseen safety-budget combinations, whereas CORMA generalizes more robustly.}
\end{figure}

Existing offline safe RL methods mainly include model-based (Kidambi et al., 2020; Yu et al., 2020), model-free (Kumar et al., 2020; Kostrikov et al., 2021), and conditional sequence modeling approaches (Chen et al., 2021; Xue et al., 2026).
Model-based and model-free methods often rely on Bellman backups, making them susceptible to overestimation and bootstrapping errors in offline settings. Moreover, they are typically trained for fixed safety budgets and thus generalize poorly to unseen safety budgets during training (Achiam et al., 2017). In contrast, methods such as the Constrained Decision Transformer (CDT) condition action generation on return-to-go (RTG) and cost-to-go (CTG), enabling zero-shot adaptation to different safety budgets through conditional sequence modeling (Liu et al., 2023b).

Despite their effectiveness, existing Transformer-based methods are primarily designed for single-constraint tasks and face two major limitations in multi-constraint settings. First, self-attention has quadratic complexity with respect to sequence length, while representing additional constraints requires more CTG tokens, substantially increasing training and inference overhead (Dao et al., 2022; Tay et al., 2022). Second, standard self-attention lacks an explicit mechanism for coordinating multiple CTG conditions and their influence on RTG-guided action generation (Vaswani et al., 2017). 

Mamba, a selective state space model (SSM), provides a promising alternative. Its parallel scan mechanism captures long-range causal dependencies with linear computational complexity, while its selective state mechanism dynamically filters task-relevant information (Gu and Dao, 2024). Mamba has shown strong performance in conditional sequence modeling for unconstrained offline RL (Lv et al., 2024; Huang et al., 2024), yet its potential for safe RL with multiple constraints remains unexplored. This motivates the following question:  \textit{How can Mamba-based conditional sequence modeling efficiently coordinate multiple safety constraints and generalize zero-shot to unseen safety-budget combinations?}

To address this, we propose \underline{\textbf{C}}\underline{\textbf{O}}nstraint-\underline{\textbf{R}}outed decision \underline{\textbf{M}}\underline{\textbf{A}}mba (CORMA), a novel Mamba-based conditional sequence modeling framework for generalizable multi-constraint offline safe RL. To be specific, CORMA introduces a constraint-aware expert-routing mechanism, where independent RTG expert networks evaluate diverse safety budgets and model feasible RTG distributions. A gating router assigns adaptive weight scores to these experts, and fuses their predicted distributions through a product-of-experts (PoE) mechanism. The final RTG is then sampled from the fused distribution and fed into a Mamba-based action decoder. To improve training stability, we design a three-stage optimization strategy that separately trains the action decoder, RTG experts, and the router. Experimental results demonstrate that CORMA achieves strong reward-safety trade-offs in multi-constraint safe RL tasks and robust zero-shot generalization to unseen safety-budget combinations. The main contributions of this paper are summarized as follows:
\begin{itemize}
\item We identify fundamental limitations of conditional sequence modeling in multi-constraint offline safe RL : additional CTG conditions substantially increase the sequence modeling overhead; while directly concatenating multiple CTG tokens 
is insufficient to achieve robust generalization under unseen safety-budget combinations.

\item We propose CORMA, a novel conditional sequence modeling framework for generalizable multi-constraint offline safe RL. CORMA uses constraint-aware RTG experts to model feasible RTG distributions under different CTG conditions, and fuses them through a router-guided PoE mechanism to generate constraint-aware RTG guidance for a Mamba-based action decoder.

\item We conduct extensive experiments on OSRL benchmarks. Results verify that CORMA achieves superior task performance while reduces inference overhead by $37\%$ compared with Transformer-based baselines, and exhibits stronger zero-shot adaptation to unseen safety-budget combinations than existing offline safe RL baselines.
\end{itemize}

\section{Preliminaries}
\subsection{Multi-Constraint Safe RL}
We model the problem of safe RL as a CMDP (Altman, 1998). A CMDP can be represented by the tuple $(\mathcal{S}, \mathcal{A}, P, r,  \{c^i\}_{i=1}^N)$, where $\mathcal{S}$ and $\mathcal{A}$ denote the state space and action space, respectively; $P: \mathcal{S} \times \mathcal{A} \times \mathcal{S} \rightarrow [0,1]$ is the state transition probability; $r: \mathcal{S} \times \mathcal{A} \rightarrow \mathbb{R}$ is the reward function; $c^i: \mathcal{S} \times \mathcal{A} \rightarrow \mathbb{R}$ represents the cost function of the $i$-th  constraint; $N$ is the number of constraints. We define the policy as $\pi: \mathcal{S} \times \mathcal{A} \to [0,1]$. The objective is to find an optimal policy $\pi^*$ that maximizes the expected cumulative reward while satisfying the constraints, i.e.,
\begin{align}
\pi^* =& \arg \max_{\pi}  \quad \mathbb{E}_{ \pi} \left[\sum_{t=1}^Tr_t\right], \\
\text{s.t.} & \quad \mathbb{E}_{\pi} \left[\sum_{t=1}^Tc^i_t\right] \leq \kappa^i ,\forall i\nonumber
\end{align}
where $\kappa^i$ represents the safety budget of the $i$-th constraint. 
\subsection{Decision Transformer For Offline Safe RL}
The traditional CDT solves the safe RL problem via a conditional sequence modeling method. It models the trajectory sequence of the agent as $\tau = (s_0, a_0, R_0,  C_0, \dots, s_T, a_T, r_T,  C_T)$, where $R_t = \sum_{j=t}^T r_j$ represents the RTG, $C_t = \sum_{j=t}^T c_j$ represents the CTG. The pre-collected dataset is defined as $\mathcal{D}$. The input of traditional CDT is $\tau_{-K:t} = (s_{t-K+1},C_{t-K+1},R_{t-K+1},a_{t-K+1},\dots,s_t,C_{t},$ $R_t)$, which captures the trajectory sequence from the previous $K$ steps. Based on this, the policy of traditional CDT is modeled as a random Gaussian distribution (Zheng et al., 2022), i.e., $a_t \sim \pi_\theta(\cdot|\tau_{-K:t}) = \mathcal{N}(\mu_\theta(\tau_{-K:t}), \Sigma_\theta(\tau_{-K:t}))$.  During the training phase, traditional CDT optimizes the policy parameters $\theta$ by minimizing the negative log-likelihood of the actions from the sampled dataset $\mathcal{D}$, regularized by a policy entropy term to improve robustness. During the inference phase, traditional CDT sets the initial guidance signals $R_1$ and $C_1$ at the first time step based on the target reward returns and safety budget $\kappa$. After executing the sampled action $a_t$, the system dynamically updates the RTG and CTG based on the instantaneous rewards $r_t$ and costs $c_t$ received from the environment, i.e., $R_{t+1} = R_t - r_t$ and $C_{t+1} = C_t - c_t$. The updated $R_{t+1}$ and $C_{t+1}$, together with the next state $s_{t+1}$, are then fed as input for auto-regressively predicting the subsequent action $a_{t+1}$.

\subsection{State Space Model and Mamba}

A continuous-time SSM describes the mapping of an input signal $x(t) \in \mathbb{R}^D$ to an output signal $y(t) \in \mathbb{R}^D$ through a hidden state $h(t) \in \mathbb{R}^N$. 
A standard linear SSM is given by
$h'(t) = \mathbf{A}h(t) + \mathbf{B}x(t)$, $y(t) = \mathbf{C}h(t)$, where $\mathbf{A} \in \mathbb{R}^{N \times N}$, $\mathbf{B} \in \mathbb{R}^{N \times D}$ and $\mathbf{C} \in \mathbb{R}^{D \times N}$ are the state transition, input projection, and output projection matrices.
To process discrete-time sequences, the continuous parameters are discretized with a step size $\Delta$. Using zero-order hold (ZOH), we obtain \(
\bar{\mathbf{A}} = \exp(\Delta \mathbf{A}),
\)\(
\bar{\mathbf{B}} = (\Delta \mathbf{A})^{-1}(\exp(\Delta \mathbf{A}) - \mathbf{I}) \cdot \Delta \mathbf{B}.
\)
The resulting discrete recurrence can be written as $h_t = \bar{\mathbf{A}}h_{t-1} + \bar{\mathbf{B}}x_t, 
y_t = \mathbf{C}h_t.$ 
Although traditional SSMs have shown promise in sequence modeling, they rely on linear time-invariant dynamics where the matrices $\mathbf{\bar{A}}$, $\mathbf{\bar{B}}$, and $\mathbf{C}$ remain fixed across time (Gu et al., 2022). This time-invariant property limits their ability to dynamically filter task-relevant information. Mamba addresses this limitation by introducing a data-dependent selection mechanism (Gu and Dao, 2023), where $\mathbf{B}$, $\mathbf{C}$, and $\Delta$ are parameterized as functions of the current input $x_t$. This enables Mamba to selectively preserve important contextual information while filtering irrelevant noise. Furthermore, by leveraging a hardware-aware parallel scan algorithm, Mamba achieves efficient long-sequence modeling with linear complexity of $O(L)$. 
\section{Investigating Conditional Sequence Modeling in Multi-Constraint Safe RL}
   In this section, we investigate two key challenges of applying conditional sequence modeling to multi-constraint safe RL: high computational complexity and limited generalization to unseen safety-budget combinations.

\subsection{Computational Complexity Challenges Under Multiple Constraints}
Traditional offline safe RL algorithms, such as Constrained Policy Optimization (CPO) (Achiam et al., 2017), and Constraints Penalized Q-learning (CPQ) (Xu et al., 2022), rely on trust regions or conservative penalties to satisfy constraints. When extended to multi-constraint scenarios via Lagrange multipliers (Tessler et al., 2018; Stooke et al., 2020; Yu et al., 2022) or separate cost critics (Guo et al., 2025), 
they suffer from gradient conflicts during shared parameter updates, since different constraints may induce inconsistent policy-update directions. This limitation motivates a sequence modeling alternative that formulates offline safe RL as trajectory generation conditioned on RTG and CTG tokens. However, the representation of multiple constraints requires additional CTG tokens. Due to the quadratic complexity of self-attention, the longer sequences cause computational costs to grow quadratically, which results in significant inference overhead.
To demonstrate this, we conducted experiments on an offline safe RL dataset derived from Safety-Gymnasium. We compared the standard CDT model with Constrained Decision Mamba (CDM), a variant that replaces the Transformer backbone with Mamba. As illustrated in Fig. 2 (a), while CDT suffers from a sharp increase in inference latency as the number of constraints increases, CDM consistently achieves lower latency. This demonstrates the better scalability of Mamba for multi-constraint offline safe RL.
\subsection{Generalization Challenges under Dynamic Combinations of Safety Budgets}
\begin{figure}[t]
    \centering
    \includegraphics[width=0.5\textwidth]{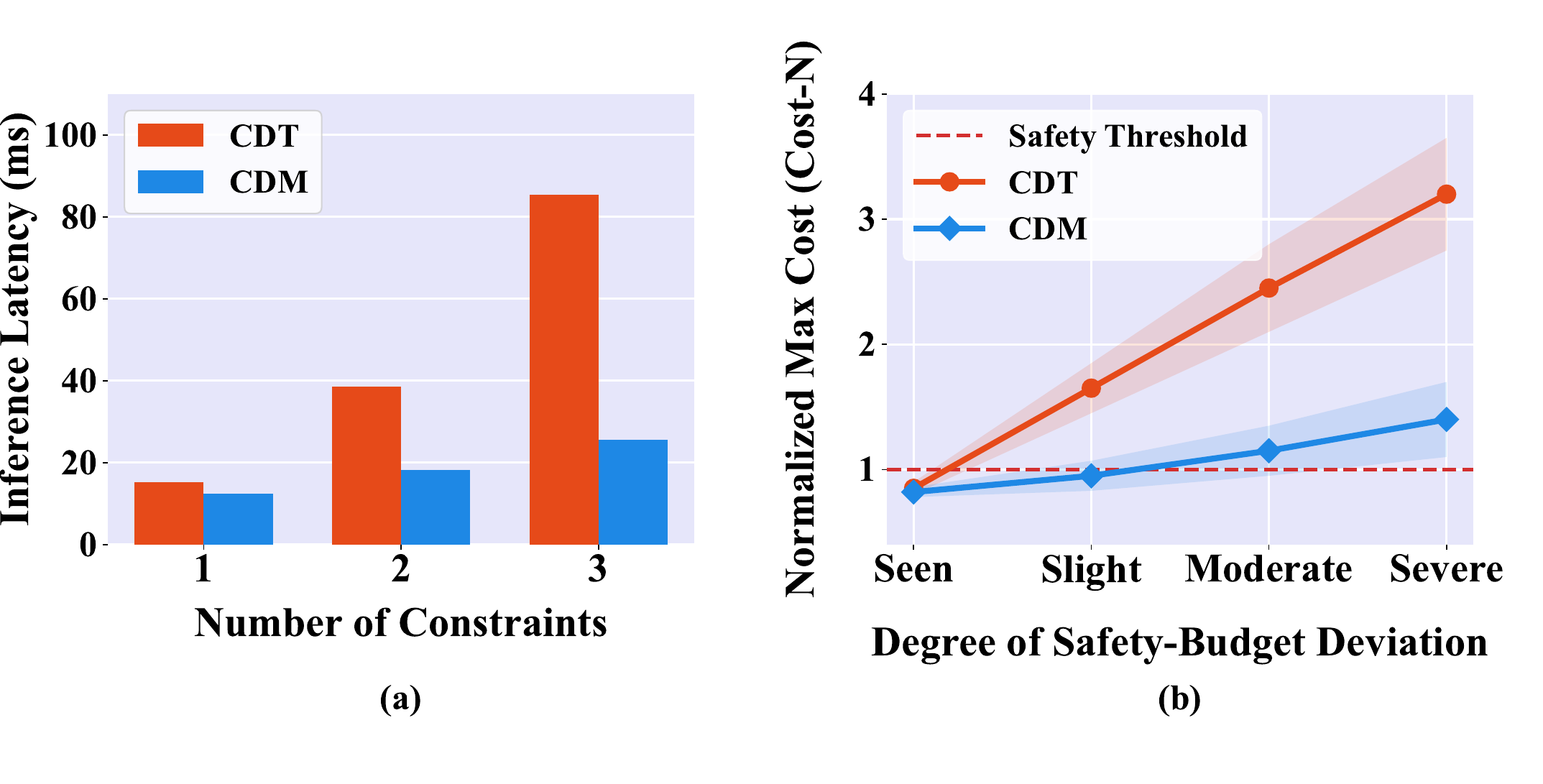} 
    \caption{Inference efficiency and zero-shot generation analysis of CDT and CDM under multi-constraint settings.
(a) Inference latency with increasing constraints. (b) Safety performance under different safety-budget regions.}

    \label{fig:2} 
\end{figure}
 Traditional Lagrangian-based methods optimize policies under static safety budgets and require full retraining when safety budgets change. Recent conditional sequence modeling methods, such as CDT (Liu et al., 2023b), SDT (Guo et al., 2024), and CoPDT (Xue et al., 2026), achieve generalization to unseen safety budgets in single-constraint environments. However, whether these methods can still generalize when multiple safety budgets change simultaneously in multi-constraint tasks remains underexplored.
To investigate this, we evaluate sequence models across safety-budget combinations with different degrees of deviation from the offline dataset. Since different constraints have heterogeneous cost scales, we normalize each cost return $C^i$ by its corresponding safety budget $\kappa^i$, and measure overall safety using Cost-$N$, defined as Cost-$N=\max_i
\{C^i/\kappa^i\}$. As visualized in Fig.~1, we partition the safety-budget space into four regions according to the minimum normalized relative distance from an evaluation budget vector $\boldsymbol{\kappa}$ to the seen region $\mathcal{K}_{\rm seen}$:
$d(\boldsymbol{\kappa},\mathcal{K}_{\rm seen})
=
\min_{\boldsymbol{\kappa}'\in\mathcal{K}_{\rm seen}}
\sqrt{
\frac{1}{N}
\sum_{i=1}^{N}
\left(
\frac{\kappa^i-\kappa'^i}{\kappa'^i}
\right)^2
}.$
The regions are seen ($d=0$), slight ($0<d\leq d_1$),
moderate ($d_1<d\leq d_2$), and severe ($d_2<d\leq d_3$), where $d_1$, $d_2$, and $d_3$ denote the predefined region
boundaries, and their specific values
are provided in appendix. As Fig. 2(b) reveals, standard CDT suffers from severe constraint violations under unseen budget combinations. Furthermore, CDM improves performance over CDT, but still fails to satisfy all constraints when multiple safety budgets change simultaneously.
These results raise a key question: \textit{How can we preserve Mamba’s efficient inference capability while achieving zero-shot generalization to unseen multi-constraint scenarios?}
\section{Method}
In this section, we first introduce the constraint-routed RTG prediction module, which models constraint-specific feasible RTG distributions and fuses them through a router-guided PoE mechanism. Then, we present the Mamba-based action decoder for conditional action generation. Finally, we describe the three-stage optimization mechanism for decoupled training of the action decoder, experts, and router.
\begin{figure*}[t]
\centering
\includegraphics[width=1\textwidth]{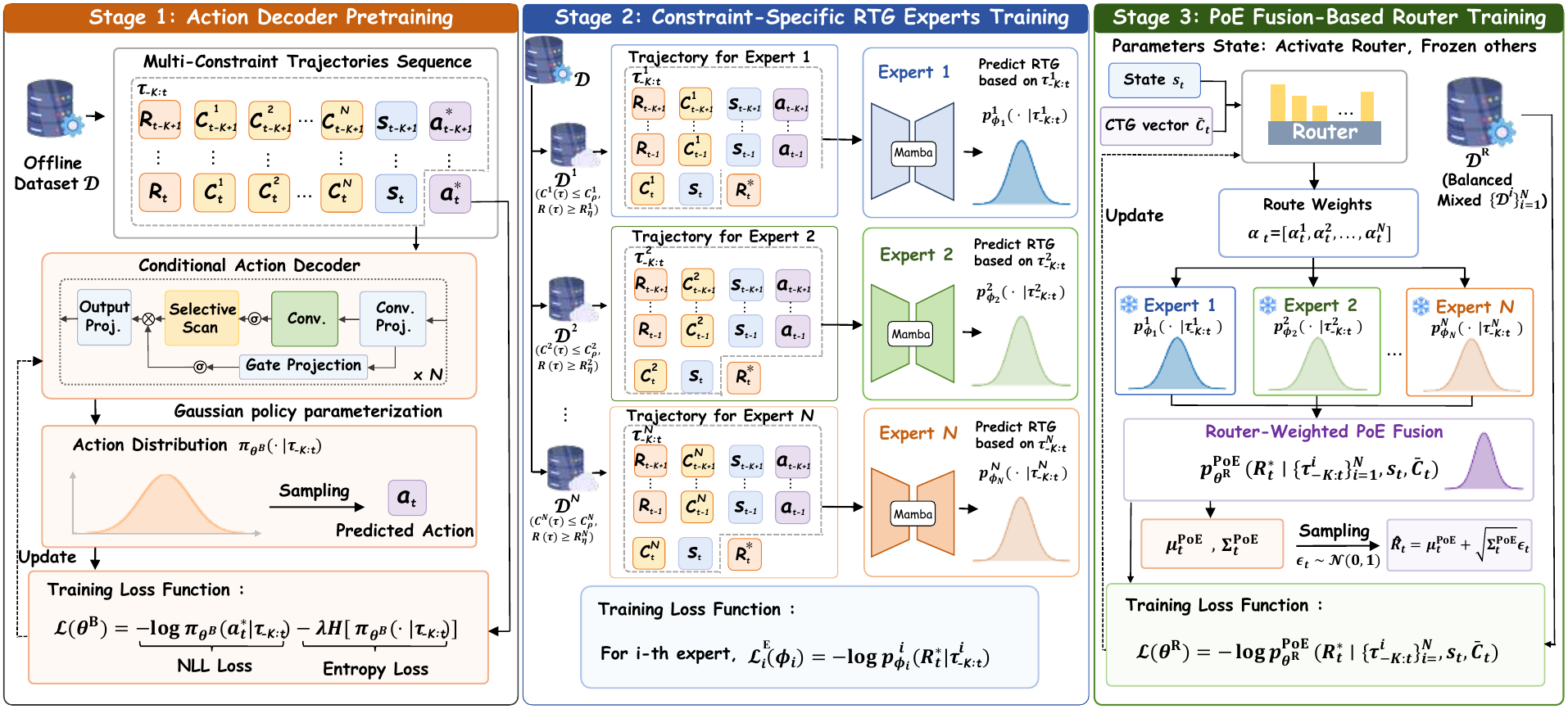}
\caption{Overview of CORMA. \textbf{Stage I} pre-trains the Mamba-based action decoder using offline trajectories. \textbf{Stage II} trains constraint-specific RTG experts to model feasible RTG distributions under different constraints. \textbf{Stage III} trains the router to adaptively fuse expert predictions through PoE and generate constraint-aware RTG guidance for safe action execution.}
\label{fig:training_framework}
\end{figure*}
\subsection{Constraint-Routed PoE for RTG Estimation}
Under the conditional sequence modeling framework, policy inference is conditioned on both target return and the remaining safety budget, which are represented by the RTG token $R_t$ and the CTG tokens $\{C^i_t\}_{i=1}^N$, respectively. Nevertheless, an inherent tension exists between $R_t$ and $C^i_t$. If an unrealistically high $R_t$ is specified, the model may over-prioritize reward maximization and consequently incur constraint violations. Thus, it is imperative to set an appropriate $R_t$ that balances reward maximization and constraint satisfaction. Moreover, in multi-constraint tasks, different constraints may require conflicting target returns. Satisfying the safety budget of one constraint may contradict that of another. This trade-off significantly reduces the feasible range of $R_t$. 
To address this, we design a constraint-routed PoE mechanism for RTG estimation. Specifically, we employ lightweight Mamba encoders as independent constraint-specific experts, where each expert models the RTG distribution for one constraint.
For the $i$-th constraint, we construct a length-$K$ historical trajectory that only contains the corresponding $C_{t}^{i}$, i.e.,
\begin{align}
\tau_{-K:t}^i=\Big(&s_{t-K+1},C_{t-K+1}^i,R_{t-K+1},a_{t-K+1},\dots,s_{t-1},\nonumber\\
&C_{t-1}^i,R_{t-1},a_{t-1},s_t,C_{t}^i\Big).
\end{align}

Based on this trajectory, the $i$-th expert parameterizes a Gaussian distribution over the current RTG, i.e.,
\begin{equation}
p_i(\cdot\mid \tau_{-K:t}^i)
=
\mathcal{N}(\mu_{\phi_i}(\tau_{-K:t}^i),\Sigma_{\phi_i}(\tau_{-K:t}^i)),
\end{equation}
where $\mu_{\phi_i}(\tau_{-K:t}^{i})$ and $\Sigma_{\phi_i}(\tau_{-K:t}^{i})$ denote the mean and the variance of the RTG distribution predicted by the $i$-th expert, respectively. Furthermore, we design a router to estimate the fusion weight for each expert's prediction. The router takes the current state $s_t$ and the normalized remaining CTG vector $\bar{\bm{C}}_t = \left[\bar{C}^1_t, \bar{C}^2_t, \dots, \bar{C}^N_t\right]^T$ as input, and outputs a normalized weight vector $\bm\alpha_t = [\alpha^1_t, \dots,  \alpha^N_t]^T$ through a softmax function, where $\bar{C}^i_t=C_t^i/\kappa^i$ denotes the normalized CTG of the $i$-th constraint, and $\alpha^i_t$ denotes the fusion weight of the $i$-th expert. Given the distributions
$\{p_i(\cdot\mid\tau_{-K:t}^i)\}_{i=1}^{N}$ and the
weights $\bm\alpha_t$, we define the fused PoE distribution as
\begin{align}
&p_{\theta^{\rm R}}^{\rm PoE}
\left(
\cdot\mid
\{\tau_{-K:t}^i\}_{i=1}^{N},
s_t,\bar{\bm C}_t
\right)\nonumber\\
&=
\frac{1}{Z_t}
\prod_{i=1}^{N}
p_i
\left(
\cdot\mid\tau_{-K:t}^i
\right)^{\alpha_t^i},
\end{align}
where $Z_t$ is the normalization constant. The final target
RTG is sampled as
\begin{equation}
\hat R_t
\sim
p_{\theta^{\rm R}}^{\rm PoE}
\left(
\cdot\mid
\{\tau_{-K:t}^i\}_{i=1}^{N},
s_t,\bar{\bm C}_t
\right).
\end{equation}
The closed-form expressions for the fused Gaussian distribution and its reparameterized sampling procedure are provided in the appendix.

\begin{algorithm}[t]
\caption{CORMA Three-Stage Training}
\label{alg:training}
\footnotesize
\setlength{\algorithmicindent}{1.0em}
\textbf{Input}: Offline dataset $\mathcal{D}$, sequence length $K$, number of constraints $N$, hyper-parameters $\lambda$, and training steps $T_1,T_2,T_3$. \

\textbf{Initialize}: $\theta^{\rm B}$, $\{\phi_i\}_{i=1}^{N}$, and $\theta^{\rm R}$.
\begin{algorithmic}[1]
\STATE \text{\textbf{// Stage 1:} Action Decoder Pre-training}
\FOR{$t=1$ to $T_1$}
\STATE Sample a batch of $(\tau_{-K:t},a^*_t)$ from $\mathcal{D}$.
\STATE Update $\theta^{\rm B}$ by minimizing $\mathcal{L}^{\rm B}(\theta^{\rm B})$ in Eq.~(9).
\ENDFOR
\STATE Freeze $\theta^{\rm B}$.

\STATE \text{\textbf{// Stage 2:} Constraint-Specific Expert Training}
\FOR{$i=1$ to $N$}
\STATE Construct $\mathcal{D}^i$ for the $i$-th expert.
\FOR{$t=1$ to $T_2$}
\STATE Sample a batch of $(\tau_{-K:t}^{i},R^*_t)$ from $\mathcal{D}_i$.
\STATE Update $\phi_i$ by minimizing $\mathcal{L}_i^{\rm E}(\phi_i)$ in Eq.~(10).
\ENDFOR
\ENDFOR
\STATE Freeze $\{\phi_i\}_{i=1}^{N}$.

\STATE \text{\textbf{// Stage 3:} Router Network Training}
\STATE Construct $\mathcal{D}^{\rm R}$ based on $\{D^i\}_{i=1}^N$.
\FOR{$t=1$ to $T_3$}
\STATE Sample a batch of $(s_t,\{C_t^i\}_{i=1}^{N},a_t,\{\tau_{-K:t}^{i}\}_{i=1}^{N})$ from $\mathcal{D}^{\rm R}$.
\STATE Compute $\bar{\bm C}_t$ and predict $\bm\alpha_t$ based on the router.
\FOR{ $i=1$ to $N$}
\STATE Predict $p_{\phi_i}(\cdot|\tau_{-K:t}^{i})$ based on the $i$-th expert network.
\ENDFOR

\STATE Sample $\hat{R}_t$ from the fused distribution \(\frac{1}{Z_t}\prod_{i=1}^{N}p_i(\cdot|\tau_{-K:t}^{i})^{\alpha_t^i}\).
\STATE Update $\theta^{\rm R}$ by minimizing $\mathcal{L}^{\rm R}(\theta^{\rm R})$ in Eq.~(11).
\ENDFOR
\STATE \textbf{Return} $\theta^{\rm B}$, $\{\phi_i\}_{i=1}^{N}$, and $\theta^{\rm R}$.
\end{algorithmic}
\end{algorithm}

\subsection{Mamba-based Conditional Action Decoder}
After obtaining the constraint-aware $\hat{R}_t$, we employ a Mamba module as the conditional action decoder. At time step $t$, the module receives a length-$K$ trajectory, i.e.,
\begin{align}
&\hat{\tau}_{-K:t}=\Big( s_{t-K+1},\{C^i_{t-K+1}\}_{i=1}^N,\hat{R}_{t-K+1},  a_{t-K+1}, \dots, \nonumber\\  
&s_{t-1},\{C^i_{t-1}\}_{i=1}^N,\hat{R}_{t-1}, a_{t-1}, s_t, \{C^i_{t}\}_{i=1}^N, \hat{R}_t \Big).
\end{align}

By leveraging its selective state space mechanism, the Mamba module extracts temporal features from this sequence to produce a hidden representation $y_t$. To improve robustness in offline policy learning, we model the action decoder as a stochastic continuous distribution. Specifically, $y_t$ is fed into two parallel linear projection heads which parameterize the mean $\mu_{\theta^{\rm B}}(\hat{\tau}_{-K:t})$ and covariance $\Sigma_{\theta^{\rm B}}(\hat{\tau}_{-K:t})$ of the action distribution, respectively. Finally, the conditional action policy is modeled as a Gaussian distribution:
\begin{equation}
\pi_{\theta^{\rm B}}(\cdot \mid \hat\tau_{-K:t}) = \mathcal{N}\Big(\mu_{\theta^{\rm B}}(\hat{\tau}_{-K:t}), \Sigma_{\theta^{\rm B}}(\hat{\tau}_{-K:t})\Big),
\end{equation}
where $\theta^{\rm B}$ denotes the parameters of the action decoder.

\subsection{Three-Stage Optimization Mechanism}
In multi-constraint offline safe RL, directly modeling multiple heterogeneous safety constraints within a unified policy optimization framework remains challenging due to the complex interactions among different constraints. To address this, we introduce a three-stage optimization mechanism, as illustrated in Fig. 3.
\subsubsection{Stage I: Conditional Action Decoder Pre-training}
In the first stage, we pre-train the conditional action decoder network via the offline dataset $\mathcal{D}$. At time step $t$, it receives a length-$K$ trajectory sequence as input, i.e., 
\begin{align}
\tau_{-K:t}=\Big(&s_{t-K+1}, \{C^i_{t-K+1}\}_{i=1}^N,R_{t-K+1},  a_{t-K+1}, \dots,  \nonumber\\
&s_t,\{C^i_{t}\}_{i=1}^N, R_t\Big).
\end{align}

To further enhance the exploration capability of the policy, we introduce a Shannon entropy regularization term $H[\pi_{\theta^{\rm B}}(\cdot |\tau_{-K:t})]$ into the optimization objective. Finally, the loss function of the network can be expressed as
\begin{align}
\mathcal{L}^{\rm B}(\theta^{\rm B})=\mathbb{E}_{(\tau_{-K:t}, a_t^*) \sim \mathcal{D}} &\Big[ -\log \pi_{\theta^{\rm B}}(a_t^* | \tau_{-K:t})\nonumber\\
&  - \lambda H\big[\pi_{\theta^{\rm B}}(\cdot | \tau_{-K:t})\big] \Big],
\end{align}
where $\lambda$ is a weight coefficient, and $a^*_t $ is the action from $\mathcal{D}$.
\subsubsection{Stage II: Constraint-Specific Expert Training}
After completing the first-stage training, we freeze the parameters $\theta^{\rm B}$ of the conditional action decoder and proceed to train each constraint-specific RTG expert. For the $i$-th constraint, let
$C_{\rho}^{i}$ denote the $\rho$-quantile of the trajectory
cost returns $\{C^i(\tau):\tau\in\mathcal D\}$. Among the trajectories
satisfying $C^i(\tau)\leq C_{\rho}^{i}$, we further define
$R_{\eta}^{i}$ as the $\eta$-quantile of their reward returns. We construct
$\mathcal D^i
=
\left\{
\tau\in\mathcal D:
C^i(\tau)\leq C_{\rho}^{i},
\;
R(\tau)\geq R_{\eta}^{i}
\right\},$
which retains low-cost trajectories with relatively high
returns under the $i$-th constraint. For each sequence
$\tau_{-K:t}^i$ sampled from $\mathcal D^i$, only the CTG
tokens associated with the $i$-th constraint are retained,
and the corresponding empirical $R_t^*$ is used as the
likelihood target of the expert. Thus, the optimization objective of the $i$-th expert network is 
\begin{equation}
\mathcal{L}^{\rm E}_{i}(\phi_i) = \mathbb{E}_{(\tau_{-K:t}^i, R_t^*) \sim \mathcal{D}^i} \Big[ -\log p_{\phi_i}(R^*_t \mid \tau_{-K:t}^i) \Big],
\end{equation}
where $\phi_i$ are the parameters of the $i$-th expert network.
\subsubsection{Stage III: Routing Network Training}
After training all experts, we freeze $\theta^{\rm B}$ and
$\{\phi_i\}_{i=1}^{N}$, and optimize only the router
parameters $\theta^{\rm R}$. We construct the routing dataset
$\mathcal D^{\rm R}$ as a balanced mixture of
$\{\mathcal D^i\}_{i=1}^{N}$ by sampling each
constraint-specific subset with equal probability. The router
is trained to maximize the likelihood of the empirical $R_t^*$
under the fused PoE distribution, i.e.,
\begin{align}
&\mathcal L^{\rm R}(\theta^{\rm R})=
\mathbb E_{
(\{\tau_{-K:t}^i\}_{i=1}^{N},s_t,\bar{\bm C}_t,R_t^*)
\sim\mathcal D^{\rm R}}
\big[-\log
p_{\theta^{\rm R}}^{\rm PoE}
\big(
R_t^*\mid
\nonumber\\
&\qquad\qquad\qquad\qquad
\{\tau_{-K:t}^i\}_{i=1}^{N},
s_t,\bar{\bm C}_t
\big)
\big].
\end{align}

\section{Experiment}

\begin{table*}[!t]
\centering
\small 
\renewcommand{\arraystretch}{1.1} 
\setlength{\tabcolsep}{2.98pt}  
\begin{tabular}{l *{10}{cc}}
\toprule
\multirow{2}{*}{\textbf{Method}}
& \multicolumn{2}{c}{\textbf{BC-v2}} & \multicolumn{2}{c}{\textbf{CC-v2}} 
& \multicolumn{2}{c}{\textbf{BC-v3}} & \multicolumn{2}{c}{\textbf{CC-v3}} & \multicolumn{2}{c}{\textbf{DC-v3}}
& \multicolumn{2}{c}{\textbf{PB-v3}} & \multicolumn{2}{c}{\textbf{PG-v3}}
& \multicolumn{2}{c}{\textbf{CG-v3}} & \multicolumn{2}{c}{\textbf{PP-v3}} & \multicolumn{2}{c}{\textbf{CP-v3}}\\
\cmidrule(lr){2-3} \cmidrule(lr){4-5} \cmidrule(lr){6-7} \cmidrule(lr){8-9} \cmidrule(lr){10-11} \cmidrule(lr){12-13} \cmidrule(lr){14-15} \cmidrule(lr){16-17} \cmidrule(lr){18-19} \cmidrule(lr){20-21}
& $r\uparrow$ & $c\downarrow$ & $r\uparrow$ & $c\downarrow$ & $r\uparrow$ & $c\downarrow$ & $r\uparrow$ & $c\downarrow$ & $r\uparrow$ & $c\downarrow$ & $r\uparrow$ & $c\downarrow$ & $r\uparrow$ & $c\downarrow$ & $r\uparrow$ & $c\downarrow$ & $r\uparrow$ & $c\downarrow$ & $r\uparrow$ & $c\downarrow$ \\
\midrule

BCQ-Lag     & 88.3 & 1.25 & \textbf{75.9} & \textbf{0.99} & 74.5 & 1.71 & 69.7 & 2.88 & 92.1 & 1.92 & 80.2 & 1.10 & 86.3 & 1.01 & 81.2 & 1.18 & 79.1 & 1.29 & 80.3 & 1.22 \\
BEAR-Lag    & 90.1 & 1.02 & 83.8 & 1.05 & 86.3 & 1.99 & 73.2 & 2.42 & 94.8 & 2.76 & 85.9 & 1.08 & 89.2 & 1.17 & 84.9 & 1.69 & 82.7 & 1.33 & 83.1 & 1.29 \\
PDCA        & \textbf{87.7} & \textbf{0.98} & \textbf{81.3} & \textbf{0.97} & 78.7 & 1.76 & 71.6 & 1.79 & 89.1 & 1.57 & 86.3 & 1.46 & 89.3 & 1.45 & 87.3 & 1.67 & 83.0 & 1.06 & 85.0 & 1.08 \\
\midrule
POCE        & 88.8 & 1.42 & 81.5 & 1.08 & 79.4 & 1.37 & 72.4 & 1.38 & 86.8 & 1.33 & {87.2} & {1.01} & \textbf{90.1} & \textbf{0.99} & 88.6 & 1.01 & 84.2 & 1.03 & 86.1 & 1.02 \\
OASIS       & \textbf{82.5} & \textbf{0.95} & 77.8 & 1.02 & 73.8 & 1.18 & 67.5 & 1.24 & 81.6 & 1.78 & \textbf{84.5} & \textbf{0.99} & \textbf{87.0} & \textbf{0.97} & \textbf{85.5} & \textbf{0.98} & 80.8 & 1.01 & \textbf{83.2} & \textbf{0.99} \\
CAPS        & \textbf{79.9} & \textbf{0.99} & \textbf{75.5} & \textbf{0.98} & 71.3 & 1.13 & \textbf{65.8} & \textbf{0.98} & 78.0 & 1.20 & \textbf{83.4} & \textbf{0.94} & 86.5 & 1.05 & \textbf{84.8} & \textbf{0.96} & \textbf{78.9} & \textbf{0.98} & \textbf{81.5} & \textbf{0.97} \\
CCAC        & \textbf{83.3} & \textbf{0.98} & \underline{\textbf{80.2}} & \underline{\textbf{0.96}} & 77.5 & 1.08 & 71.2 & 1.34 & \textbf{84.2} & \textbf{0.99} & 88.3 & 1.16 & \textbf{90.5} & \textbf{0.94} & \textbf{88.9} & \textbf{0.95} & \textbf{83.8} & \textbf{0.96} & \textbf{86.2} & \textbf{0.99} \\
\midrule
BC-Safe     & 75.9 & 1.98 & 77.2 & 1.32 & 82.4 & 1.14 & 72.6 & 1.65 & \textbf{43.0} & \textbf{0.98} & 70.4 & 1.59 & 58.5 & 1.77 & 61.2 & 1.04 & 81.1 & 1.25 & 70.8 & 1.13 \\
CDT-MC      & \textbf{72.8} & \textbf{0.97} & 75.1 & 1.21 & 78.2 & 1.62 & 71.8 & 1.08 & 86.2 & 1.35 & 70.3 & 1.91 & 92.2 & 1.47 & 70.4 & 1.22 & 77.5 & 1.23 & 79.1 & 1.65 \\
CoPDT-MC    & 89.9 & 1.61 & \textbf{79.6} & \textbf{0.99} & 79.8 & 1.49 & 71.3 & 1.39 & 87.3 & 1.73 & 88.6 & 1.69 & 88.5 & 1.37 & 87.4 & 1.27 & 86.4 & 1.23 & 87.9 & 1.91 \\
CORDT       & \textbf{83.5} & \textbf{0.91} & 82.4 & 1.16 & 80.2 & 1.28 & \underline{\textbf{72.4}} & \underline{\textbf{0.92}} & \textbf{83.5} & \textbf{0.96} & \textbf{88.2} & \textbf{0.97} & {90.9} &{1.02} & \textbf{86.2} & \textbf{0.95} & \textbf{85.5} & \textbf{0.98} & 87.1 & 1.01 \\
\midrule
CORMA w/o PoE& 70.4 & 1.18 & \textbf{62.8} & \textbf{0.99} & 75.8 & 1.48 & \textbf{71.1} & \textbf{0.99} & 81.9 & 1.18 & 65.0 & 1.15 & \textbf{70.1} & \textbf{0.97} & 68.0 & 1.16 & 79.4 & 1.05 & \textbf{54.0} & \textbf{0.98} \\
CORMA-S     & 82.9 & 1.19 & 78.4 & 1.23 & 73.8 & 1.87 & 66.8 & 1.69 & 80.1 & 1.78 & 75.3 & 1.39 & 79.7 & 1.68 & 78.1 & 1.27 & 74.6 & 1.36 & 79.7 & 1.29 \\
\midrule
\textbf{CORMA (Ours)} & \underline{\textbf{84.8}} & \underline{\textbf{0.89}} & \textbf{78.4} & \textbf{0.98} & \underline{\textbf{80.9}} & \underline{\textbf{0.89}} & \textbf{70.8} & \textbf{0.97} & \underline{\textbf{88.0}} & \underline{\textbf{0.95}} & \underline{\textbf{90.2}} & \underline{\textbf{0.93}} & \underline{\textbf{91.7}} & \underline{\textbf{0.92}} & \underline{\textbf{89.5}} & \underline{\textbf{0.94}} & \underline{\textbf{86.9}} & \underline{\textbf{0.93}} & \underline{\textbf{88.3}} & \underline{\textbf{0.95}}\\
\bottomrule
\end{tabular}
\caption{Evaluation results on Bullet-Safety-Gym and Safety-Gymnasium tasks. Entries report normalized reward ($r\uparrow \in [0, 100]$) and maximum normalized cost among all constraints ($c\downarrow$), averaged over 5 seeds and 20 episodes. \textbf{Black bold} indicates constraint satisfaction ($c \leq 1$). \underline{\textbf{Black bold}} highlights the best safety-performance trade-off among all methods.}
\label{tab:main_results_cross_benchmark}
\end{table*}
In this section, we design our experiments to answer the following research questions:
\begin{itemize} 
\item RQ1: Can CORMA learn safe policies under multiple constraints from offline datasets?
\item RQ2: Can CORMA achieve both few-shot and zero-shot adaptation to unseen safety-budget combinations?
\item RQ3: Can CORMA achieve efficient training and deployment under multi-constraint settings?
\end{itemize}
\subsection{Experimental Setup}

\begin{figure*}[t]
    \centering
    \includegraphics[width=1.0\textwidth, keepaspectratio]{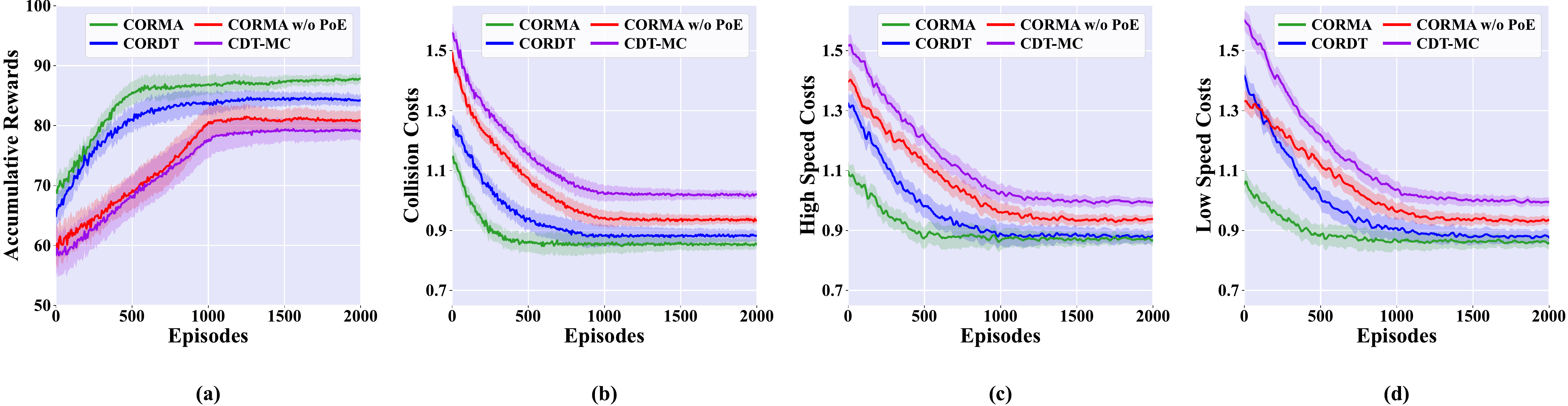}
    \caption{
    Fine-tuning convergence under new safety-budget combinations in the DC-v3 tasks. 
    The policies are fine-tuned using trajectories that satisfy the new safety-budget combinations. 
    (a) Normalized accumulative rewards during fine-tuning. 
    (b) Normalized collision costs. 
    (c) Normalized high-speed costs. 
    (d) Normalized low-speed costs.
      }
    \label{fig:fine_tuning_convergence}
\end{figure*}
\begin{figure*}[t]
    \centering
    \includegraphics[width=0.98\textwidth, keepaspectratio]{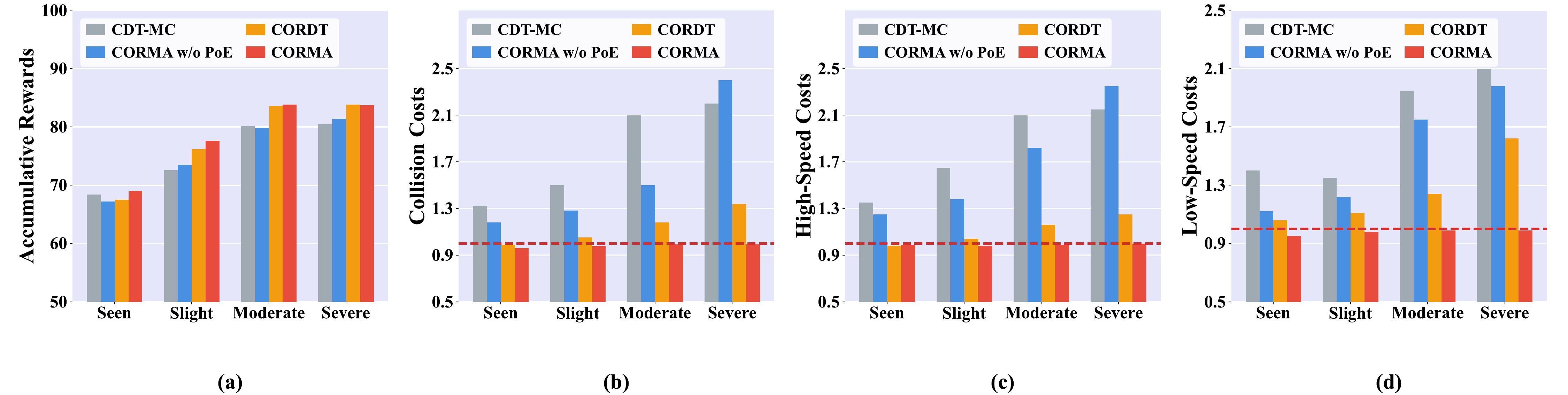}
    \caption{
    Zero-shot generalization under unseen safety-budget combinations in the DC-v3 task. 
    (a) Normalized accumulative rewards. 
    (b) Normalized Collision costs. 
    (c) Normalized High-speed costs. 
    (d) Normalized Low-speed costs.
     }
    \label{fig:zero_shot_dc_v3}
\end{figure*}
\subsubsection{Tasks and Constraints}
We evaluate our method on OSRL benchmark tasks built upon Bullet-Safety-Gym (Liu et al., 2023a) and Safety-Gymnasium (Ji et al., 2023). We consider four types of robots (Point, Car, Ball, and Drone) and four tasks (Circle, Goal, Push, Button). Task names follow the convention ``Robot-Task", e.g., PG denotes ``Point-Goal".
For each task, we incorporate three heterogeneous constraints: the boundary/collision constraint, the high-velocity constraint, and the low-velocity constraint. Based on these, we establish two task settings: the -v2 tasks include the two constraints of boundary/collision and high velocity, while the challenging -v3 tasks encompass all constraints.  

\subsubsection{Metrics}
We evaluate performance using episodic reward return and episodic cost return. 
To ensure comparability across tasks with different reward scales, we adopt the normalized reward return as $R_{\rm normalized} = \frac{R_{\pi} - R_{\min}} {R_{\max} - R_{\min}} \times 100$,  where \(R_{\pi}\) denotes the evaluated episodic reward return of policy \(\pi\), and \(R_{\min}\) and \(R_{\max}\) are the minimum and maximum episodic reward returns in the offline dataset, respectively. 
To compare heterogeneous constraints and categorize safety-budget combinations for zero-shot evaluation, we use Cost-$N$ to measure overall safety and $d(\boldsymbol{\kappa},\mathcal{K}_{\rm seen})$ to quantify budget deviation, as predefined in the preceding
generalization analysis. Specific experimental conditions and hyperparameters are detailed in appendix.
\subsubsection{Algorithms and Baselines}
 To evaluate the performance of CORMA, we compared it against the following baselines: (1) CDT-MC, a multi-constraint adaptation of the CDT; (2) CORMA w/o PoE, an ablation baseline that completely removes the constraint-specific experts and the routing mechanism, directly feeding the concatenated CTGs and RTG into the action decoder; (3) CORMA-S, which replaces the multi-expert PoE module with a single unified network that jointly models all CTGs for RTG prediction; (4) Constraint-Routed Decision Transformer (CORDT), which substitutes the action decoder network of CORMA with Transformers; (5) CoPDT-MC, a multi-constraint adaptation of the recent sequence modeling method CoPDT (Xue et al., 2026). (6) BC-Safe (Nguyen et al., 2022), a behavior cloning baseline; (7) BCQ (Fujimoto et al., 2019) with Lagrangian (BCQ-Lag), BEAR (Kumar et al., 2019) with Lagrangian (BEAR-Lag), and PDCA (Hong et al., 2024), safe RL methods adapted for multiple constraints through Lagrangian multipliers; (8) POCE (Guan, 2024), OASIS (Yao, 2024), CAPS (Chemingui et al., 2025), CCAC (Guo et al., 2025), four recent baselines that demonstrates excellent performance; 
\subsection{Comparison and Analysis}
\subsubsection{RQ1:}
\begin{table}[!t]
\centering
\small 
\renewcommand{\arraystretch}{0.9} 
\setlength{\tabcolsep}{3pt} 
\begin{tabular*}{\columnwidth}{@{\extracolsep{\fill}}lcccc}
\toprule
Phase & CDT & CDM & CORDT & CORMA \\
\midrule
Backbone (h)    & 15.82 & 11.64 & 15.81 & 11.62 \\
Experts (h)     & --    & --    & 1.42  & 1.34  \\
Router (h)      & --    & --    & 0.71  & 0.68  \\
\midrule
Total Train (h) & 15.82 & 11.64 & 17.94 & \textbf{13.64} \\
Inference (s)   & 0.008 & 0.003 & 0.010 & \textbf{0.005} \\
\bottomrule
\end{tabular*}
\caption{Time comparison during training and deployment. Training times are reported in hours (h), and deployment inference times are reported in seconds (s).}
\label{tab:time_complexity} 
\end{table}
As shown in Table~\ref{tab:main_results_cross_benchmark}, CORMA achieves a favorable reward--safety trade-off across most tasks. Conventional offline safe RL methods remain competitive on \text{-v2} tasks, but degrade markedly on the challenging \text{-v3} tasks. More importantly, CORDT consistently outperforms CORMA w/o PoE, indicating that the constraint-routed PoE mechanism is the primary source of CORMA's performance gains. Meanwhile, CORMA-S frequently violates constraints, indicating that a single expert is insufficient for reliable multi-constraint RTG prediction. CORMA w/o PoE generally outperforms CDT-MC and CoPDT-MC, confirming the complementary benefit of the Mamba action decoder. These results demonstrate that CORMA can effectively learn safety-aware policies from offline datasets under multiple heterogeneous constraints.
\subsubsection{RQ2:}
To evaluate CORMA's adaptability, we first examine its
few-shot adaptation to unseen safety-budget combinations on the
challenging DC-v3 task. The models are pre-trained on the base
dataset and then fine-tuned using a small subset of
trajectories. As shown in Fig.~4, CORMA reduces the collision, high-speed, and low-speed
costs below their budgets with fewer fine-tuning steps.
The overall performance follows
CORMA $>$ CORDT $>$ CORMA w/o PoE $>$ CDT-MC.
The advantage of CORDT over CDT-MC indicates the effectiveness of constraint-routed PoE mechanism, while the gain of CORMA over CORDT shows that the Mamba action decoder improves
adaptability.
Moreover, we evaluate the zero-shot generalization of CORMA under slight, moderate, and severe safety-budget regions, corresponding to increasing
values of
$d(\boldsymbol{\kappa},\mathcal{K}_{\rm seen})$.
As shown in Fig.~5, CORMA maintains competitive rewards across different regions. Fig.~5(b)-(d) further show that CORMA controls all individual costs within their
budgets, whereas the baselines exhibit increasing violations. These results demonstrate the robust zero-shot adaptability of CORMA
to unseen safety-budget combinations.
\subsubsection{RQ3:}
We evaluate the computational efficiency of CORMA during training and deployment. Although CORMA adds additional experts and a router, the linear complexity of  Mamba significantly reduces sequence modeling overhead compared with Transformer-based baselines. As shown in Table 2, this computational efficiency advantage offsets the overhead of the additional modules, leading to lower training time than the Transformer-based baselines. Moreover, CORMA achieves 0.005 s/step inference latency, demonstrating its ability to support multi-constraint policy inference with lower deployment latency.

\section{Conclusion}
In this work, we study multi-constraint offline safe RL with varying safety-budget combinations. We propose CORMA, a framework that decomposes policy inference into constraint-aware RTG estimation and adaptive action generation. A three-stage optimization mechanism separately trains the action decoder, experts, and router. Experiments on OSRL benchmarks show that CORMA achieves better reward-safety trade-offs and stronger zero-shot generalization to unseen budget combinations than existing safe RL baselines.

\section*{References}

\nobibliography{aaai2027}
\bibentry{99}\\[.2em]
\bibentry{89}\\[.2em]
\bibentry{107}\\[.2em]
\bibentry{124}\\[.2em]
\bibentry{97}\\[.2em]
\bibentry{101}\\[.2em]
\bibentry{120}\\[.2em]
\bibentry{87}\\[.2em]
\bibentry{90}\\[.2em]
\bibentry{109}\\[.2em]
\bibentry{104}\\[.2em]
\bibentry{123}\\[.2em]
\bibentry{114}\\[.2em]
\bibentry{127}\\[.2em]
\bibentry{83}\\[.2em]
\bibentry{88}\\[.2em]
\bibentry{122}\\[.2em]
\bibentry{106}\\[.2em]
\bibentry{118}\\[.2em]
\bibentry{93}\\[.2em]
\bibentry{96}\\[.2em]
\bibentry{121}\\[.2em]
\bibentry{91}\\[.2em]
\bibentry{98}\\[.2em]
\bibentry{105}\\[.2em]
\bibentry{119}\\[.2em]
\bibentry{112}\\[.2em]
\bibentry{102}\\[.2em]
\bibentry{111}\\[.2em]
\bibentry{103}\\[.2em]
\bibentry{86}\\[.2em]
\bibentry{84}\\[.2em]
\bibentry{85}\\[.2em]
\bibentry{110}\\[.2em]
\bibentry{95}\\[.2em]
\bibentry{100}\\[.2em]
\bibentry{126}\\[.2em]
\bibentry{113}\\[.2em]
\bibentry{94}\\[.2em]
\bibentry{92}\\[.2em]
\bibentry{108}\\[.2em]

\clearpage
\onecolumn
\appendix
\begin{adjustwidth}{2cm}{2cm}
\section*{Appendix}
\vspace{1.0cm}
\noindent\textbf{A. Gaussian Representation of the Router-Weighted PoE} \dotfill 11 \\[1.5ex]
\noindent\textbf{B. Generalization Bound of CORMA under Multi-Constraint Offline RL} \dotfill 12 \\[1.5ex]
\noindent\textbf{C. Related Work} \dotfill 16 \\[1.5ex]
\noindent\hspace*{2em} C.1 Offline Safe Reinforcement Learning \dotfill 16 \\[1.5ex]
\noindent\hspace*{2em} C.2 Generalization in Offline Safe RL \dotfill 17 \\[1.5ex]
\noindent\hspace*{2em} C.3 Mamba in Reinforcement Learning \dotfill 17 \\[1.5ex]
\noindent\textbf{D. CORMA Three-Stage Training and deployment algorithms} \dotfill 17 \\[1.5ex]
\noindent\textbf{E. Experimental Details} \dotfill 19 \\[1.5ex]
\noindent\hspace*{2em} E.1 Experimental Tasks and Datasets \dotfill 19 \\[1.5ex]
\noindent\hspace*{2em} E.2 Baselines \dotfill 20 \\[1.5ex]
\noindent\hspace*{2em} E.3 Safety-Budget Shift Regions Settings \dotfill 21 \\[1.5ex]
\noindent\hspace*{2em} E.4 Hyperparameters \dotfill 22 \\[1.5ex]
\noindent\textbf{F. More Experimental Results} \dotfill 22 \\[1.5ex]
\noindent\hspace*{2em} F.1 Analysis on Safe Policy Learning under Multiple Constraints \dotfill 22 \\[1.5ex]
\noindent\hspace*{2em} F.2 Analysis on Trajectory Stitching Ability \dotfill 26 \\[1.5ex]
\noindent\hspace*{2em} F.3 Analysis of the Zero-Shot Constraints Generalization \dotfill 27 \\[1.5ex]
\noindent\hspace*{2em} F.4 Effect of Trajectory Context Length on Stage-1 Action Decoder Training \dotfill 28 \\[1.5ex]
\noindent\hspace*{2em} F.5 How Critical Is the Three-Stage Training Strategy? \dotfill 30 \\[1.5ex]
\clearpage
\section*{A. Gaussian Representation of the
Router-Weighted PoE}
\label{app:poe_gaussian}
In this section, we derive the Gaussian
representation of the router-weighted
PoE distribution introduced in the main text.
For the $i$-th expert, the predicted
distribution over the RTG at time step $t$ is
$p_i
\left(
\cdot
\mid
\tau_{-K:t}^{i}
\right)
=
\mathcal N
\left(
\mu_{i,t},
\Sigma_{i,t}
\right),
\label{eq:app_expert_gaussian}$
where
$\mu_{i,t}
=
\mu_{\phi_i}
\left(
\tau_{-K:t}^{i}
\right),
\Sigma_{i,t}
=
\Sigma_{\phi_i}
\left(
\tau_{-K:t}^{i}
\right)
\label{eq:app_expert_moments}$
denote the mean and variance predicted by the $i$-th expert, respectively. Conditioned on
$\{\tau_{-K:t}^{i}\}_{i=1}^{N}$, $s_t$,
and $\bar{\bm C}_t$, the router-weighted PoE distribution is defined as
\begin{equation}
p_{\theta^{\rm R}}^{\rm PoE}
\left(
\cdot
\mid
\{\tau_{-K:t}^{i}\}_{i=1}^{N},
s_t,
\bar{\bm C}_t
\right)
=
\frac{1}{Z_t}
\prod_{i=1}^{N}
p_i
\left(
\cdot
\mid
\tau_{-K:t}^{i}
\right)^{\alpha_t^i},
\label{{eq:app_complete_square}}
\end{equation}
where $Z_t$ denotes the normalization constant. The routing
weights are generated by a softmax function and satisfy
$\alpha_t^i>0,
\sum_{i=1}^{N}\alpha_t^i=1.
\label{eq:app_router_normalization}$
Let $\zeta_t$ denote an arbitrary candidate value of the
scalar RTG at time step $t$. Substituting the Gaussian
density into the
unnormalized PoE, we have
\begin{equation}
\begin{aligned}
&\prod_{i=1}^{N}
p_i
\left(
\zeta_t
\mid
\tau_{-K:t}^{i}
\right)^{\alpha_t^i}
\\
&=
\prod_{i=1}^{N}
\left[
\frac{1}{\sqrt{2\pi\Sigma_{i,t}}}
\exp
\left(
-\frac{
(\zeta_t-\mu_{i,t})^2
}{
2\Sigma_{i,t}
}
\right)
\right]^{\alpha_t^i}
\\
&=
\left[
\prod_{i=1}^{N}
(2\pi\Sigma_{i,t})^{-\alpha_t^i/2}
\right]
\exp
\left[
-\frac{1}{2}
\sum_{i=1}^{N}
\frac{\alpha_t^i}{\Sigma_{i,t}}
(\zeta_t-\mu_{i,t})^2
\right].
\end{aligned}
\label{eq:app_gaussian_product}
\end{equation}

Define the fused precision and the corresponding
precision-weighted mean term as
\begin{equation}
\Lambda_t^{\rm PoE}
=
\sum_{i=1}^{N}
\frac{\alpha_t^i}{\Sigma_{i,t}},
\qquad
\eta_t^{\rm PoE}
=
\sum_{i=1}^{N}
\frac{\alpha_t^i\mu_{i,t}}{\Sigma_{i,t}}.
\label{eq:app_poe_natural_parameters}
\end{equation}
Since $\alpha_t^i>0$ and $\Sigma_{i,t}>0$,
$\Lambda_t^{\rm PoE}>0$. Expanding the quadratic term in
Eq.~\eqref{eq:app_gaussian_product}, we have
\begin{equation}
\begin{aligned}
&\sum_{i=1}^{N}
\frac{\alpha_t^i}{\Sigma_{i,t}}
(\zeta_t-\mu_{i,t})^2
\\
&=
\Lambda_t^{\rm PoE}\zeta_t^2
-
2\eta_t^{\rm PoE}\zeta_t
+
\sum_{i=1}^{N}
\frac{\alpha_t^i\mu_{i,t}^2}{\Sigma_{i,t}}.
\\
&=
\Lambda_t^{\rm PoE}
\left(
\zeta_t
-
\frac{\eta_t^{\rm PoE}}{\Lambda_t^{\rm PoE}}
\right)^2
+
\sum_{i=1}^{N}
\frac{\alpha_t^i\mu_{i,t}^2}{\Sigma_{i,t}}
-
\frac{
(\eta_t^{\rm PoE})^2
}{
\Lambda_t^{\rm PoE}
}.
\end{aligned}
\label{eq:app_quadratic_expansion}
\end{equation}

By definition, the normalization constant $Z_t$ ensures that the resulting probability density function integrates to one. It is given by the integral of the unnormalized PoE distribution over all possible values of $\zeta_t$, i.e.,
\begin{equation}
Z_t = \int \prod_{i=1}^{N} p_i \left( \zeta_t \mid \tau_{-K:t}^{i} \right)^{\alpha_t^i} \,\mathrm{d}\zeta_t.
\label{eq:app_zt_integral}
\end{equation}

Substituting the expanded quadratic form from Eq.~(4) into Eq.~\eqref{eq:app_gaussian_product}, the terms independent of $\zeta_t$ can be factored out of the integral. Evaluating the remaining standard Gaussian integral $\int \exp(-\frac{\Lambda}{2}x^2)\,\mathrm{d}x = \sqrt{2\pi/\Lambda}$, we obtain the explicit closed-form expression for the normalization constant, i.e.,
\begin{equation}
Z_t = \sqrt{ \frac{2\pi}{\Lambda_t^{\rm PoE}} } \left[ \prod_{i=1}^{N} (2\pi\Sigma_{i,t})^{-\alpha_t^i/2} \right] \exp \left[ -\frac{1}{2} \left( \sum_{i=1}^{N} \frac{\alpha_t^i\mu_{i,t}^2}{\Sigma_{i,t}} - \frac{(\eta_t^{\rm PoE})^2}{\Lambda_t^{\rm PoE}} \right) \right].
\label{eq:app_zt_closed_form}
\end{equation}

Dividing the unnormalized density in Eq.~\eqref{eq:app_gaussian_product} by $Z_t$, the leading constant multipliers and the exponential remainder exactly cancel out. The normalized fused density can therefore be written as
\begin{equation}
\begin{aligned}
&p_{\theta^{\rm R}}^{\rm PoE}
\left(
\zeta_t
\mid
\{\tau_{-K:t}^{i}\}_{i=1}^{N},
s_t,
\bar{\bm C}_t
\right)
\\
&=
\sqrt{
\frac{
\Lambda_t^{\rm PoE}
}{
2\pi
}
}
\exp
\left[
-\frac{
\Lambda_t^{\rm PoE}
}{2}
\left(
\zeta_t
-
\frac{
\eta_t^{\rm PoE}
}{
\Lambda_t^{\rm PoE}
}
\right)^2
\right].
\end{aligned}
\label{eq:app_normalized_poe}
\end{equation}

Hence, the router-weighted PoE remains Gaussian, i.e.,
\begin{equation}
\begin{aligned}
p_{\theta^{\rm R}}^{\rm PoE}
\left(
\cdot
\mid
\{\tau_{-K:t}^{i}\}_{i=1}^{N},
s_t,
\bar{\bm C}_t
\right)=
\mathcal N
\left(
\mu_t^{\rm PoE},
\Sigma_t^{\rm PoE}
\right),
\end{aligned}
\label{eq:app_poe_gaussian}
\end{equation}
where the fused mean and variance are
\begin{equation}
\mu_t^{\rm PoE}
=\frac{\eta_t^{\rm PoE}}{\Lambda_t^{\rm PoE}}
, \qquad \Sigma_t^{\rm PoE}
=
\frac{1}{\Lambda_t^{\rm PoE}}.
\label{eq:app_poe_mean}
\end{equation}

Finally, the target RTG is sampled from the fused
distribution 
$p_{\theta^{\rm R}}^{\rm PoE}
\left(
\cdot
\mid
\{\tau_{-K:t}^{i}\}_{i=1}^{N},
s_t,
\bar{\bm C}_t
\right)$.
The sampling
operation can be implemented through the reparameterization technique, i.e.,
\begin{equation}
\hat R_t
=
\mu_t^{\rm PoE}
+
\sqrt{
\Sigma_t^{\rm PoE}
}
\epsilon_t,
\qquad
\epsilon_t
\sim
\mathcal N(0,1).
\label{eq:app_poe_reparameterization}
\end{equation}

\section*{B. Generalization Bound of CORMA under Multi-Constraint Offline RL}

To analyze the generalization capabilities of CORMA under novel safety-budget combinations, we derive total variation distance (TVD)-based bounds on the expected cumulative return and per-constraint cumulative cost discrepancies. Based on the theoretical analysis conclusions, we further explain how the constraint-specific RTG experts and router-weighted PoE fusion reduce RTG-induced distribution shift, thereby improving multi-constraint generalization.

\textbf{Assumption 1.}
Let $\mathcal{C}_t=\{C_t^1,C_t^2,\dots,C_t^N\}$ denote the CTG collection associated with $N$ constraints. At each time step, the instantaneous reward $r(s_t,a_t)$ and cost $c^i(s_t,a_t)$ of the $i$-th constraint are bounded. There exist $R_{\max}$ and $\{C_{\max}^i\}_{i=1}^{N}$ such that
\begin{equation}
|r(s_t,a_t)| \le R_{\max},
\end{equation}
\begin{equation}
|c^i(s_t,a_t)| \le C_{\max}^i,
\quad \forall i\in\{1,\dots,N\}.
\end{equation}

\textbf{Assumption 2.}
For each constraint dimension $i\in\{1,\dots,N\}$, the maximum TVD discrepancy $\mathcal{D}_{TV}^{C^i}(t)$ between the conditional distribution of the next-step CTG generated by CORMA during online generalization
$p_1(C_{t+1}^{i}|s_{t+1},s_t,R_t,\mathcal{C}_t,a_t,C_{t+1}^{1:i-1})$, 
and the corresponding distribution under the feasible optimal policy
$p_2(C_{t+1}^{i}|s_{t+1},s_t,R_t,\mathcal{C}_t,a_t,C_{t+1}^{1:i-1})$ satisfies
\begin{align}
\max_t\,
\mathbb{E}_{\substack{
(s_t,R_t,\mathcal{C}_t)
\sim p_1^t(s_t,R_t,\mathcal{C}_t),\;
a_t\sim\pi_1(\cdot|s_t,R_t,\mathcal{C}_t),\\
(s_{t+1},C_{t+1}^{1:i-1})
\sim
p_1(\cdot
|s_t,R_t,\mathcal{C}_t,a_t)
}}
\left[
\mathcal{D}_{TV}^{C^i}(t)
\right]
\le
\xi_{C^i},
\quad
\forall i\in\{1,\dots,N\}.
\end{align}
where $\xi_{C^i}$ denotes the upper bound on the expected TVD discrepancy between the online generated conditional distribution and that of the feasible optimal policy for the $i$-th CTG.

\textbf{Assumption 3.}
The TVD discrepancy $\mathcal{D}_{TV}^{R}(t)$ between the online RTG conditional distribution generated by CORMA $p_1(R_{t+1}
|\mathcal{C}_{t+1},s_{t+1},s_t,R_t,\mathcal{C}_t,a_t)$ and the corresponding RTG conditional distribution under the feasible optimal policy $p_2(R_{t+1}
|\mathcal{C}_{t+1},s_{t+1},s_t,R_t,\mathcal{C}_t,a_t)$ satisfies
\begin{align}
\max_t\,
\mathbb{E}_{\substack{
(s_t,R_t,\mathcal{C}_t)
\sim p_1^t(s_t,R_t,\mathcal{C}_t),\;
a_t\sim\pi_1(\cdot|s_t,R_t,\mathcal{C}_t),\\
(s_{t+1},\mathcal{C}_{t+1})
\sim
p_1(\cdot
|s_t,R_t,\mathcal{C}_t,a_t)
}}
\left[
\mathcal{D}_{TV}^{R}(t)
\right]
\le
\xi_R.
\end{align}
where $\xi_R$ denotes the upper bound on the expected TVD discrepancy between the online generated RTG conditional distribution and the RTG conditional distribution of the feasible optimal policy.

\textbf{Assumption 4.}
The maximum TVD discrepancy between the online generalization policy $\pi_1$ learned by CORMA and the feasible optimal policy $\pi_2$ is bounded, i.e.,
\begin{equation}
\max_{s_t,R_t,\mathcal{C}_t}
\mathcal{D}_{TV}
\Big(
\pi_1(\cdot|s_t,R_t,\mathcal{C}_t)
\parallel 
\pi_2(\cdot|s_t,R_t,\mathcal{C}_t)
\Big)
\le
\xi_\pi,
\end{equation}
where $\xi_\pi$ denotes the upper bound on the policy discrepancy between the online generalization policy and the feasible optimal policy.

\textbf{Theorem 1}: The expected cumulative reward $\Lambda_1^R$ and the expected cumulative cost of the $i$-th constraint $\Lambda_1^{C^{i}}$ in the online deployment phase are bounded as
\begin{align}
|\Lambda_{1}^R-\Lambda_{2}^R|
&\le
2R_{\max}
\left[
T\left(\xi_0+\xi_{\pi}\right)
+
\frac{T(T-1)}{2}
\left(
\sum_{j=1}^{N}\xi_{C^{j}}
+\xi_R+\xi_{\pi}
\right)
\right],
\end{align}
\begin{align}
|\Lambda_{1}^{C^{i}}-\Lambda_{2}^{C^{i}}|
&\le
2C_{\max}^{i}
\left[
T\left(\xi_0+\xi_{\pi}\right)
+
\frac{T(T-1)}{2}
\left(
\sum_{j=1}^{N}\xi_{C^{j}}
+\xi_R+\xi_{\pi}
\right)
\right],
\quad
\forall i.
\end{align}
where $\Lambda_2^R$ and $\Lambda_2^{C^{i}}$ denote the expected cumulative reward and the expected cumulative cost of the $i$-th constraint under the feasible optimal policy, respectively; $T$ is the finite deployment horizon, and $\xi_0$ denotes the TVD discrepancy between the initial distributions.

Proof:
We regard the target RTG $R_t$ and the CTG collection $\mathcal{C}_t=\{C_t^1,C_t^2,\dots,C_t^N\}$ as part of the observation space. 
Let 
$p_2(s_{t+1},R_{t+1},\mathcal{C}_{t+1}|s_t,R_t,\mathcal{C}_t,a_t)$ 
denote the joint transition distribution under the feasible optimal policy, and let 
$p_1(s_{t+1},R_{t+1},\mathcal{C}_{t+1}|s_t,R_t,\mathcal{C}_t,a_t)$ 
denote the joint transition distribution generated by CORMA during the online generalization. 
By applying Bayes' rule and the probability chain rule, the joint transition distribution during online generalization can be factorized as
\begin{align}
&p_1(s_{t+1}, R_{t+1}, \mathcal{C}_{t+1}
|s_t, R_t, \mathcal{C}_t, a_t) \nonumber\\
&=
p_1(s_{t+1}|s_t, R_t, \mathcal{C}_t, a_t) 
\cdot\prod_{i=1}^{N}
p_1(C_{t+1}^{i}
|s_{t+1}, s_t, R_t, \mathcal{C}_t, a_t, C_{t+1}^{1:i-1}) \nonumber\\
&\quad \cdot
p_1(R_{t+1}
|\mathcal{C}_{t+1}, s_{t+1}, s_t, R_t, \mathcal{C}_t, a_t).
\end{align}
Similarly, the joint transition distribution under the feasible optimal policy can be factorized as
\begin{align}
&p_2(s_{t+1}, R_{t+1}, \mathcal{C}_{t+1}
|s_t, R_t, \mathcal{C}_t, a_t) \nonumber\\
&=
p_2(s_{t+1}|s_t, R_t, \mathcal{C}_t, a_t)  \cdot
\prod_{i=1}^{N}
p_2(C_{t+1}^{i}
|s_{t+1}, s_t, R_t, \mathcal{C}_t, a_t, C_{t+1}^{1:i-1}) \nonumber\\
&\quad \cdot
p_2(R_{t+1}
|\mathcal{C}_{t+1}, s_{t+1}, s_t, R_t, \mathcal{C}_t, a_t).
\end{align}
Since RTG and CTGs do not influence the physical state transition dynamics, we have
\begin{equation}
p_1(s_{t+1}|s_t, R_t, \mathcal{C}_t, a_t)
=
p_1(s_{t+1}|s_t,a_t),
\end{equation}
\begin{equation}
p_2(s_{t+1}|s_t, R_t, \mathcal{C}_t, a_t)
=
p_2(s_{t+1}|s_t,a_t).
\end{equation}
Moreover, since $p_1(s_{t+1}|s_t,a_t)
=
p_2(s_{t+1}|s_t,a_t)$, we have
\begin{equation}
p_1(s_{t+1}|s_t, R_t, \mathcal{C}_t, a_t)
=
p_2(s_{t+1}|s_t, R_t, \mathcal{C}_t, a_t).
\end{equation}
Therefore, we expand the TVD between the joint transition distribution during online generalization 
and the joint transition distribution under the feasible optimal policy as
\begin{align}
    & \mathcal{D}_{TV}\Big(
    p_1(s_{t+1}, R_{t+1}, \mathcal{C}_{t+1}
    |s_t, R_t, \mathcal{C}_t, a_t) \parallel 
    p_2(s_{t+1}, R_{t+1}, \mathcal{C}_{t+1}
    |s_t, R_t, \mathcal{C}_t, a_t)
    \Big) \nonumber \\
    &\le 
    \sum_{i=1}^{N} 
    \mathbb{E}_{s_{t+1}, C_{t+1}^{1:i-1} \sim p_1(\cdot|s_t,R_t,\mathcal{C}_t,a_t)} 
    \bigg[
    \mathcal{D}_{TV}\Big( 
    p_1(C_{t+1}^{i}
    |s_{t+1}, s_t, R_t, \mathcal{C}_t, a_t, C_{t+1}^{1:i-1}) \nonumber \\
    &\qquad\qquad\qquad \parallel 
    p_2(C_{t+1}^{i}
    |s_{t+1}, s_t, R_t, \mathcal{C}_t, a_t, C_{t+1}^{1:i-1})
    \Big) 
    \bigg] \nonumber \\
    &\quad + 
    \mathbb{E}_{s_{t+1}, \mathcal{C}_{t+1} \sim p_1(\cdot|s_t,R_t,\mathcal{C}_t,a_t)} 
    \bigg[
    \mathcal{D}_{TV}\Big( 
    p_1(R_{t+1}
    |\mathcal{C}_{t+1}, s_{t+1}, s_t, R_t, \mathcal{C}_t, a_t) \nonumber \\
    &\qquad\qquad\qquad \parallel 
    p_2(R_{t+1}
    |\mathcal{C}_{t+1}, s_{t+1}, s_t, R_t, \mathcal{C}_t, a_t)
    \Big) 
    \bigg].
\label{eq:inline_tvd_multi}
\end{align}
Based on \textbf{Assumptions 2} and \textbf{3}, the maximum one-step discrepancy $\xi_m$ between the joint transition distribution during online generalization and that under the feasible optimal policy satisfies
\begin{align}
\xi_m
&=
\max_t
\mathbb{E}_{\substack{
(s_t,R_t,\mathcal{C}_t)
\sim p_1^t(s_t,R_t,\mathcal{C}_t),\\
a_t\sim\pi_1(\cdot|s_t,R_t,\mathcal{C}_t)
}}
\Bigg[
\mathcal{D}_{TV}
\Big(
p_1(s_{t+1},R_{t+1},\mathcal{C}_{t+1}
|s_t,R_t,\mathcal{C}_t,a_t)
\nonumber\\
&\hspace{5.3cm}\parallel
p_2(s_{t+1},R_{t+1},\mathcal{C}_{t+1}
|s_t,R_t,\mathcal{C}_t,a_t)
\Big)
\Bigg]
\nonumber\\
&\le
\sum_{j=1}^{N}
\max_t
\mathbb{E}_{\substack{
(s_t,R_t,\mathcal{C}_t)
\sim p_1^t(s_t,R_t,\mathcal{C}_t),
a_t\sim\pi_1(\cdot|s_t,R_t,\mathcal{C}_t),\\
(s_{t+1},C_{t+1}^{1:j-1})
\sim
p_1(s_{t+1},C_{t+1}^{1:j-1}
|s_t,R_t,\mathcal{C}_t,a_t)
}}
\left[
\mathcal{D}_{TV}^{C^j}(t)
\right]
\nonumber\\
&\quad+
\max_t
\mathbb{E}_{\substack{
(s_t,R_t,\mathcal{C}_t)
\sim p_1^t(s_t,R_t,\mathcal{C}_t),
a_t\sim\pi_1(\cdot|s_t,R_t,\mathcal{C}_t),\\
(s_{t+1},\mathcal{C}_{t+1})
\sim
p_1(s_{t+1},\mathcal{C}_{t+1}
|s_t,R_t,\mathcal{C}_t,a_t)
}}
\left[
\mathcal{D}_{TV}^{R}(t)
\right]
\nonumber\\
&\le
\sum_{j=1}^{N}\xi_{C^j}+\xi_R.
\end{align}

Based on the $R_{\rm max}$ and $\{C^i_{\rm max}\}_{i=1}^N$ in \textbf{Assumption 1}, we expand the difference between the expected cumulative returns of the policies $\pi_1$ and $\pi_2$ over a finite horizon $T$ as
\begin{align}
    &|\Lambda_{1}^R-\Lambda_{2}^R| \nonumber\\
    &=
    \bigg|
    \sum_{t=0}^{T-1}
    \sum_{s_t, R_t, \mathcal{C}_t,a_t}
    \big(
    p_{1}^{t}(s_t, R_t, \mathcal{C}_t,a_t)
    -
    p_{2}^{t}(s_t, R_t, \mathcal{C}_t,a_t)
    \big)
    r(s_t, a_t)
    \bigg| \nonumber\\
    &\le 
    \sum_{t=0}^{T-1}
    \sum_{s_t, R_t, \mathcal{C}_t,a_t}
    \big|
    p_{1}^{t}(s_t, R_t, \mathcal{C}_t,a_t)
    -
    p_{2}^{t}(s_t, R_t, \mathcal{C}_t,a_t)
    \big|
    |r(s_t, a_t)| \nonumber\\
    &\le 
    R_{\max}
    \sum_{t=0}^{T-1}
    \sum_{s_t, R_t, \mathcal{C}_t,a_t}
    \big|
    p_{1}^{t}(s_t, R_t, \mathcal{C}_t,a_t)
    -
    p_{2}^{t}(s_t, R_t, \mathcal{C}_t,a_t)
    \big| \nonumber\\
    &= 
    R_{\max}
    \sum_{t=0}^{T-1}
    \sum_{s_t, R_t, \mathcal{C}_t,a_t}
    \big|
    p_{1}^{t}(s_t, R_t, \mathcal{C}_t)
    \pi_1(a_t|s_t, R_t, \mathcal{C}_t)-
    p_{2}^{t}(s_t, R_t, \mathcal{C}_t)
    \pi_2(a_t|s_t, R_t, \mathcal{C}_t)
    \big| \nonumber\\
    &=
    R_{\max}
    \sum_{t=0}^{T-1}
    \sum_{s_t, R_t, \mathcal{C}_t,a_t}
    \big|
    p_{1}^{t}(s_t, R_t, \mathcal{C}_t)
    \pi_1(a_t|s_t, R_t, \mathcal{C}_t)-
    p_{1}^{t}(s_t, R_t, \mathcal{C}_t)
    \pi_2(a_t|s_t, R_t, \mathcal{C}_t)
    \nonumber\\
    &\quad +
    \big(
    p_{1}^{t}(s_t, R_t, \mathcal{C}_t)
    -
    p_{2}^{t}(s_t, R_t, \mathcal{C}_t)
    \big)
    \pi_2(a_t|s_t, R_t, \mathcal{C}_t)
    \big| \nonumber\\
    &\le 
    R_{\max}
    \sum_{t=0}^{T-1}
    \sum_{s_t, R_t, \mathcal{C}_t,a_t}
    \big|
    p_{1}^{t}(s_t, R_t, \mathcal{C}_t)
    \pi_1(a_t|s_t, R_t, \mathcal{C}_t)
     -
    p_{1}^{t}(s_t, R_t, \mathcal{C}_t)
    \pi_2(a_t|s_t, R_t, \mathcal{C}_t)
    \big| \nonumber\\
    &\quad +
    R_{\max}
    \sum_{t=0}^{T-1}
    \sum_{s_t, R_t, \mathcal{C}_t}
    \big|
    p_{1}^{t}(s_t, R_t, \mathcal{C}_t)
    -
    p_{2}^{t}(s_t, R_t, \mathcal{C}_t)
    \big| \nonumber\\
    &= 
    2R_{\max}
    \sum_{t=0}^{T-1}
    \mathbb{E}_{(s_t, R_t, \mathcal{C}_t)\sim p_{1}^{t}}
    \big[
    \mathcal{D}_{TV}
    (
    \pi_{1}(\cdot|s_t, R_t, \mathcal{C}_t)
    \parallel
    \pi_{2}(\cdot|s_t, R_t, \mathcal{C}_t)
    )
    \big] \nonumber\\
    &\quad +
    2R_{\max}
    \sum_{t=0}^{T-1}
    \mathcal{D}_{TV}
    (
    p_{1}^{t}(s_t, R_t, \mathcal{C}_t)
    \parallel
    p_{2}^{t}(s_t, R_t, \mathcal{C}_t)
    ),
\end{align}
where $\Lambda_{1}^R$ and $\Lambda_{2}^R$ denote the expected cumulative returns of $\pi_1$ and $\pi_2$, respectively.

Furthermore, based on the policy discrepancy bound $\xi_\pi$ in \textbf{Assumption 4}, we have 
\begin{align}
    &|\Lambda_{1}^R-\Lambda_{2}^R| \nonumber\\
    &\le 
    2R_{\max}
    \sum_{t=0}^{T-1}
    \xi_{\pi}
     +
    2R_{\max}
    \sum_{t=0}^{T-1}
    \mathcal{D}_{TV}
    \Big(
    p_{1}^{t}(s_t, R_t, \mathcal{C}_t)
    \parallel 
    p_{2}^{t}(s_t, R_t, \mathcal{C}_t)
    \Big).
\end{align}

Let $\xi_t
=
\mathcal{D}_{TV}
\big(
p_{1}^{t}(s_t, R_t, \mathcal{C}_t)
\parallel
p_{2}^{t}(s_t, R_t, \mathcal{C}_t)
\big).$ In particular, the initial distribution discrepancy is defined as
\begin{align}
\xi_0
=
\mathcal{D}_{TV}
\Big(
p_{1}^{0}(s_0,R_0,\mathcal{C}_0)
\parallel
p_{2}^{0}(s_0,R_0,\mathcal{C}_0)
\Big).
\end{align} We recursively expand this discrepancy based on the one-step Markov transition from time step $t-1$ to $t$ as
\begin{align}
    \xi_t
    &=
    \mathcal{D}_{TV}
    \Big(
    p_{1}^{t}(s_t, R_t, \mathcal{C}_t)
    \parallel 
    p_{2}^{t}(s_t, R_t, \mathcal{C}_t)
    \Big)
    \nonumber\\
    &=
    \frac{1}{2}
    \sum_{s_t, R_t, \mathcal{C}_t}
    \big|
    p_{1}^{t}(s_t, R_t, \mathcal{C}_t)
    -
    p_{2}^{t}(s_t, R_t, \mathcal{C}_t)
    \big|
    \nonumber\\
    &\le
    \frac{1}{2}
    \mathbb{E}_{(s_{t-1}, R_{t-1}, \mathcal{C}_{t-1})
    \sim p_{1}^{t-1}(s_{t-1}, R_{t-1}, \mathcal{C}_{t-1})}
    \bigg[
    \sum_{s_t, R_t, \mathcal{C}_t}
    \big|
    p_{1}(s_t, R_t, \mathcal{C}_t
    |s_{t-1}, R_{t-1}, \mathcal{C}_{t-1})
    \nonumber\\
    &\quad -
    p_{2}(s_t, R_t, \mathcal{C}_t
    |s_{t-1}, R_{t-1}, \mathcal{C}_{t-1})
    \big|
    \bigg]
    \nonumber\\
    &\quad +
    \mathcal{D}_{TV}
    \Big(
    p_{1}^{t-1}(s_{t-1}, R_{t-1}, \mathcal{C}_{t-1})
    \parallel
    p_{2}^{t-1}(s_{t-1}, R_{t-1}, \mathcal{C}_{t-1})
    \Big).
\end{align}

We define the one-step discrepancy between the two joint-state transition distributions as
\begin{align}
\delta_t
=&
\frac{1}{2}
\mathbb{E}_{(s_{t-1},R_{t-1},\mathcal{C}_{t-1})
\sim p_1^{t-1}(s_{t-1},R_{t-1},\mathcal{C}_{t-1})}
\bigg[
\sum_{s_t, R_t, \mathcal{C}_t}
\big|
p_{1}(s_t, R_t, \mathcal{C}_t
|s_{t-1}, R_{t-1}, \mathcal{C}_{t-1})\nonumber\\
&-
p_{2}(s_t, R_t, \mathcal{C}_t
|s_{t-1}, R_{t-1}, \mathcal{C}_{t-1})
\big|
\bigg].
\end{align}

We further decompose the joint-state transition into the action distribution induced by the policy and the environment transition dynamics. For $p_1$, we have
\begin{align}
    &p_1(s_t, R_t, \mathcal{C}_t
    |s_{t-1}, R_{t-1}, \mathcal{C}_{t-1})
    \nonumber\\
    &= 
    \sum_{a_{t-1}} 
    \pi_1(a_{t-1}
    |s_{t-1}, R_{t-1}, \mathcal{C}_{t-1})\cdot 
    p_1(s_t, R_t, \mathcal{C}_t
    |s_{t-1}, R_{t-1}, \mathcal{C}_{t-1}, a_{t-1}).
\end{align}

Similarly, for $p_2$, we have
\begin{align}
    &p_2(s_t, R_t, \mathcal{C}_t
    |s_{t-1}, R_{t-1}, \mathcal{C}_{t-1})
    \nonumber\\
    &= 
    \sum_{a_{t-1}} 
    \pi_2(a_{t-1}
    |s_{t-1}, R_{t-1}, \mathcal{C}_{t-1})\cdot 
    p_2(s_t, R_t, \mathcal{C}_t
    |s_{t-1}, R_{t-1}, \mathcal{C}_{t-1}, a_{t-1}).
\end{align}

Therefore, by applying the triangle inequality, we obtain
\begin{align}
    \delta_t 
    &= 
    \frac{1}{2} 
    \mathbb{E}_{(s_{t-1}, R_{t-1}, \mathcal{C}_{t-1}) 
    \sim p_1^{t-1}(s_{t-1}, R_{t-1}, \mathcal{C}_{t-1})}
    \Bigg[
    \sum_{s_t, R_t, \mathcal{C}_t}
    \bigg|
    \sum_{a_{t-1}}
    \Big[
    \pi_1(a_{t-1}
    |s_{t-1}, R_{t-1}, \mathcal{C}_{t-1})
    \nonumber\\
    &\quad \cdot 
    p_1(s_t, R_t, \mathcal{C}_t
    |s_{t-1}, R_{t-1}, \mathcal{C}_{t-1}, a_{t-1})
     -
    \pi_2(a_{t-1}
    |s_{t-1}, R_{t-1}, \mathcal{C}_{t-1})
    \nonumber\\
    &\quad \cdot
    p_2(s_t, R_t, \mathcal{C}_t
    |s_{t-1}, R_{t-1}, \mathcal{C}_{t-1}, a_{t-1})
    \Big]
    \bigg|
    \Bigg]\nonumber\\
    &\le 
    \mathbb{E}_{(s_{t-1}, R_{t-1}, \mathcal{C}_{t-1}) 
    \sim p_1^{t-1}(s_{t-1}, R_{t-1}, \mathcal{C}_{t-1})}
    \Bigg[
    \sum_{a_{t-1}} 
    \pi_1(a_{t-1}
    |s_{t-1}, R_{t-1}, \mathcal{C}_{t-1})
    \nonumber\\
    &\quad \cdot 
    \mathcal{D}_{TV}
    \Big(
    p_1(s_t, R_t, \mathcal{C}_t
    |s_{t-1}, R_{t-1}, \mathcal{C}_{t-1}, a_{t-1})
    \parallel 
    p_2(s_t, R_t, \mathcal{C}_t
    |s_{t-1}, R_{t-1}, \mathcal{C}_{t-1}, a_{t-1})
    \Big)
    \Bigg]
    \nonumber\\
    &\quad + 
    \mathbb{E}_{(s_{t-1}, R_{t-1}, \mathcal{C}_{t-1}) 
    \sim p_1^{t-1}(s_{t-1}, R_{t-1}, \mathcal{C}_{t-1})}
    \Bigg[
    \mathcal{D}_{TV}
    \Big(
    \pi_1(\cdot
    |s_{t-1}, R_{t-1}, \mathcal{C}_{t-1})
    \parallel 
    \pi_2(\cdot
    |s_{t-1}, R_{t-1}, \mathcal{C}_{t-1})
    \Big)
    \Bigg]\nonumber
\end{align}
\begin{align}
    &\le 
    \xi_m
     +
    \max_{s_{t-1}, R_{t-1}, \mathcal{C}_{t-1}}
    \mathcal{D}_{TV}
    \Big(
    \pi_1(\cdot
    |s_{t-1}, R_{t-1}, \mathcal{C}_{t-1})
    \nonumber\\
    &\qquad \parallel 
    \pi_2(\cdot
    |s_{t-1}, R_{t-1}, \mathcal{C}_{t-1})
    \Big)
    \nonumber\\
    &\le 
    \xi_m + \xi_\pi
    \nonumber\\
    &\le 
    \sum_{i=1}^{N}\xi_{C^i} + \xi_R + \xi_\pi.
\end{align}

Substituting this into the recursion of the marginal distribution discrepancy, we have
\begin{align}
    \xi_t
    &=
    \mathcal{D}_{TV}
    \Big(
    p_{1}^{t}(s_t, R_t, \mathcal{C}_t) 
    \parallel 
    p_{2}^{t}(s_t, R_t, \mathcal{C}_t)
    \Big)
    \nonumber\\
    &\le 
    \delta_t+\xi_{t-1}
    \nonumber\\
    &\le 
    \xi_0+\sum_{i=1}^{t}\delta_i
    \nonumber\\
    &\le 
    \xi_0+t
    \Big(
    \sum_{i=1}^{N} \xi_{C^{i}} + \xi_R + \xi_{\pi}
    \Big).
\end{align}

Finally, substituting the marginal distribution discrepancy bound $\xi_t \le \xi_0+t \big( \sum_{i=1}^{N} \xi_{C^{i}} + \xi_R + \xi_{\pi} \big)$, we obtain the following upper bound on the expected return discrepancy, i.e.,
\begin{align}
|\Lambda_{1}^R-\Lambda_{2}^R|
&\le
2R_{\max}
\sum_{t=0}^{T-1}
\left[
\xi_0
+
t
\left(
\sum_{j=1}^{N}\xi_{C^j}
+\xi_R+\xi_\pi
\right)
+\xi_\pi
\right]
\nonumber\\
&=
2R_{\max}
\left[
T\left(\xi_0+\xi_\pi\right)
+
\frac{T(T-1)}{2}
\left(
\sum_{j=1}^{N}\xi_{C^j}
+\xi_R+\xi_\pi
\right)
\right].
\end{align}

Similarly, for any constraint dimension $i \in \{1,\dots,N\}$, by replacing the reward function with the $i$-th cost function and scaling the bound by the maximum one-step cost $C_{\max}^{i}$, we obtain the upper bound on the expected cumulative cost discrepancy as
\begin{align}
|\Lambda_{1}^{C^i}-\Lambda_{2}^{C^i}|
&\le
2C_{\max}^{i}
\sum_{t=0}^{T-1}
\left[
\xi_0
+
t
\left(
\sum_{j=1}^{N}\xi_{C^j}
+\xi_R+\xi_\pi
\right)
+\xi_\pi
\right]
\nonumber\\
&=
2C_{\max}^{i}
\left[
T\left(\xi_0+\xi_\pi\right)
+
\frac{T(T-1)}{2}
\left(
\sum_{j=1}^{N}\xi_{C^j}
+\xi_R+\xi_\pi
\right)
\right],
\quad
\forall i.
\end{align}

\textbf{Theorem 1} characterizes the performance gap between the online deployment policy and the feasible optimal policy under new budget combinations. This gap is primarily determined by the policy discrepancy $\xi_{\pi}$, the RTG transition discrepancy $\xi_{R}$, and the CTG transition discrepancies $\{\xi_{C^i}\}_{i=1}^{N}$. When confronted with unseen safety-budget combinations, an idealized feasible optimal policy can allocate an appropriate target return RTG based on the current state and remaining budgets. However, when handling such unseen constraint conditions, traditional CDT-based methods fail to generate suitable RTGs, leading to a substantial RTG transition discrepancy $\xi_R$ against the feasible optimal policy, which compounds rapidly over time steps. In contrast, by thoroughly learning the intrinsic relationships among states, CTGs, and RTGs from offline expert trajectories, CORMA can dynamically set well-calibrated RTGs when encountering unseen budget combinations. This effectively reduces the RTG transition discrepancy $\xi_R$ against the feasible optimal policy, thereby tightening the performance-gap bounds and enabling the cumulative return $\Lambda_1^R$ and per-constraint cumulative cost $\Lambda_1^{C^i}$ to more closely approach the idealized feasible optimal performance.

\section*{C. Related Work}
\subsection{C.1 Offline Safe Reinforcement Learning}
Offline safe reinforcement learning learns constraint-satisfying policies from static datasets and avoids risky online exploration in real-world environments. Early methods mainly address single-cost constraints. For example, CPO enforces safety by constraining each policy update within a trust-region optimization framework (Achiam et al., 2017). CBPL extends safe policy learning to offline batch settings by combining fitted policy evaluation with constrained policy improvement (Le et al., 2019). CPQ further reduces offline constraint violations by assigning conservative penalties to unsafe or out-of-distribution actions during Q-learning (Xu et al., 2022). However, many practical scenarios require agents to satisfy multiple safety constraints simultaneously. To address the complex challenges of multi-constraint settings, recent studies have developed specific gradient manipulation techniques. For instance, (Yao et al. , 2022) approach multi-constraint safe RL from a Multi-Objective Optimization (MOO) perspective, proposing a Gradient Shaping (GradS) method for Lagrangian-based algorithms that improves training efficiency by eliminating redundant and conflicting constraint gradients. Similarly, (Kim et al. 2023) introduce a trust region-based Safe Distributional Actor-Critic (SDAC) algorithm, which features a gradient integration method that manages infeasibility issues by projecting unsafe policies into a feasible policy set through a quadratic program consisting of all constraint gradients. Nevertheless, due to the inherent trade-offs among different constraints, these methods may still suffer from policy generalization under unseen safety-budget combinations.

\subsection{C.2 Generalization in Offline Safe RL}
Recent studies have explored generalization in offline safe RL. For instance, CCAC enables zero-shot adaptation to varying constraint thresholds by modeling the relationship between state-action distributions and safety constraints to regularize critics and policy learning (Guo et al., 2025). CDT improves adaptation to varying constraint thresholds by conditioning action generation on target returns and cost returns (Liu et al., 2023b). CoPDT advances multi-task safety adaptation by using a constraint-prioritized prompt encoder to distinguish different constraints and a dynamic RTG generation mechanism to predict target returns aligned with current safety requirements (Xue et al., 2026). However, existing methods still mainly generalize to threshold changes under a single constraint. They remain insufficient for multi-constraint settings where several safety thresholds vary simultaneously.

\subsection{C.3 Mamba in Reinforcement Learning}

Mamba is a selective state space model that enables linear-time
sequence modeling through an input-dependent selection mechanism and
a hardware-aware parallel scan (Gu and Dao, 2024). Its efficiency in
processing long trajectories has motivated several applications in
reinforcement learning. Ota et al. replace the self-attention modules
of Decision Transformer with Mamba for sequential decision-making
(Ota, 2024). Lv et al. further propose a multi-grained Decision Mamba
that captures both inter-step temporal dependencies and intra-step
relationships among RTGs, states, and actions, achieving improved
performance on MuJoCo tasks (Lv et al., 2024).
Subsequent studies explore different applications of Mamba in
trajectory modeling. Huang et al. introduce Decision Mamba-Hybrid,
which combines Mamba-based long-term memory with Transformer-based
action prediction and achieves faster inference on
long-horizon tasks (Huang et al., 2024). Dai et al. show that a
carefully designed Decision Mamba can outperform Decision Transformer
with substantially fewer parameters on Atari and MuJoCo
(Dai et al., 2024). Nevertheless, existing Mamba-based
RL methods mainly focus on unconstrained return maximization and
computational efficiency. To the best of our knowledge, Mamba-based
conditional sequence modeling has not yet been studied for offline safe RL under varying and unseen
safety-budget combinations.
\section*{D. CORMA Three-Stage Training and deployment algorithms}

In this section, we provide the detailed algorithms of CORMA in multi-constraint scenarios. As described in the main paper, the workflow of CORMA primarily includes two phases: the three-stage offline training and the online policy deployment.

The offline training phase consists of three decoupled stages. In Stage I, we pretrain the Mamba-based action decoder backbone $\theta^{\rm B}$. We sample length-$K$ multi-constraint trajectories $\tau_{-K:t}$ from the offline dataset $\mathcal{D}$. The backbone learns the conditional action policy $\pi_{\theta^{\rm B}}$ via minimization of the loss function $\mathcal{L}^{\rm B}(\theta^{\rm B})$. After this stage, the backbone parameters $\theta^{\rm B}$ are frozen. In Stage II, we train the constraint-specific experts. For each constraint $i$, we filter the dataset to obtain a specific subset $\mathcal{D}^i$. The $i$-th expert network $\phi_i$ processes the constraint-specific trajectory sequence $\tau_{-K:t}^i$ to predict the feasible RTG distribution $p_{\phi_i}(\cdot|\tau_{-K:t}^i)$. These experts are optimized through minimization of the objective $\mathcal{L}_i^{\rm E}(\phi_i)$ and subsequently frozen. In Stage III, we train the gating router $\theta^{\rm R}$ on a subset $\mathcal{D}^{\rm R}$. The router evaluates the current state $s_t$ and the normalized remaining safety budget vector $\bar{\bm{C}}_t$ to compute the routing weights $\bm\alpha_t$. The model fuses the expert predictions into a product of experts distribution to sample the final target RTG $R_t$. The router updates its parameters via minimization of the loss $\mathcal{L}^{\rm R}(\theta^{\rm R})$. The detailed three-stage training procedure is provided in Algorithm 1.

During the online deployment phase, the well-trained networks coordinate to generate safe actions. At each time step $t$, the agent observes the current state $s_t$ and appends the state and CTGs to the trajectory sequences for each expert. Each expert infers its respective RTG distribution based on the historical context. The router then evaluates the normalized safety budgets $\bar{\bm{C}}_t$ to assign the weights $\bm\alpha_t$. These weights fuse the expert distributions to sample a constraint-aware target RTG $\hat{R}_t$. Finally, the Mamba backbone $\theta^{\rm B}$ utilizes the updated full trajectory sequence $\hat{\tau}_{-K:t}$ as the condition to sample an action $a_t$. After the execution of $a_t$, the environment returns the instant reward and costs to update the remaining safety budgets $C_{t+1}^i$. The deployment procedure is provided in Algorithm 2.

\begin{minipage}{0.8\textwidth}
\begin{algorithm}[H]
\caption{CORMA Three-Stage Training}
\label{alg:training}
\textbf{Input}: Offline dataset $\mathcal{D}$, sequence length $K$, number of constraints $N$, hyper-parameters $\lambda, \rho, \eta$, action decoder pretraining steps $T_1$, expert training steps $T_2$, router training steps $T_3$. \\
\textbf{Initialize}: action decoder $\theta^{\rm B}$, experts $\{\phi_i\}_{i=1}^N$, router $\theta^{\rm R}$.
\begin{algorithmic}[1]
\STATE {// Stage 1: Action Decoder Pretraining}
\FOR{$t=1$ to $T_1$}
    \STATE Sample a batch of length-$K$ trajectories $\tau_{-K:t} = (s_{t-K+1},R_{t-K+1}, \{C^i_{t-K+1}\}_{i=1}^N,  a_{t-K+1},$ $\dots, s_t,R_{t}, \{C^i_{t}\}_{i=1}^N)$ and  actions $a^*_t$ from $\mathcal{D}$.
    \STATE Predict the conditional action policy $\pi_{\theta^{\rm B}}(\cdot|\tau_{-K:t})$.
    \STATE Update Mamba backbone $\theta^{\rm B}$ by minimizing $\mathcal{L}^{\rm B}(\theta^{\rm B})$.
\ENDFOR
\STATE Freeze parameters $\theta^{\rm B}$.

\STATE {// Stage 2: Constraint-Specific Expert Training}
\FOR{each constraint $i \in \{1, \dots, N\}$}
    \STATE Filter dataset to obtain $\mathcal{D}^i$ for the $i$-th expert.
    \FOR{$t=1$ to $T_2$}
        \STATE Sample a batch of constraint-specific trajectories $\tau_{-K:t}^i$ and  RTG $R^*_t$ from $\mathcal{D}^i$.
        \STATE Predict RTG distribution $p_{\phi_i}(\cdot|\tau_{-K:t}^i)$ based on the $i$-th expert network.
        \STATE Update expert $\phi_i$ by minimizing $\mathcal{L}_i^{\rm E}(\phi_i)$.
    \ENDFOR
\ENDFOR
\STATE Freeze all expert parameters $\{\phi_i\}_{i=1}^N$.

\STATE {// Stage 3: Router Network Training}
\STATE Construct $\mathcal{D}^{\rm R}$ based on $\{D^i\}_{i=1}^N$.
\FOR{$t=1$ to $T_3$}
    \STATE Sample a batch of $(s_t, \{C^i_t\}_{i=1}^N, R^*_t)$ and the trajectory $\{\tau_{-K:t}^i\}_{i=1}^N$from $\mathcal{D}^{\rm R}$.
    \STATE Compute normalized remaining safety budgets vector $\bm{\bar{C}}_t$.
    \STATE Predict routing weights $\bm\alpha_t = [\alpha_t^1, \dots, \alpha_t^N]^T$ based on the router.
    \FOR{each constraint $i \in \{1, \dots, N\}$}
    \STATE Predict RTG distribution $p_{\phi_i}(\cdot|\tau_{-K:t}^i)$ based on $i$-th expert network.
    \ENDFOR
    \STATE Sample the final target RTG $\hat{R}_t$ from the fused PoE distribution $\frac{1}{Z_t} \prod_{i=1}^{N} p_i(\cdot\mid \tau_{-K:t}^i)^{\alpha_t^i}$.
    \STATE Update router $\theta^{\rm R}$ by minimizing $\mathcal{L}^{\rm R}(\theta^{\rm R})$.
\ENDFOR
\STATE \textbf{Return} Well-trained models $\theta^{\rm B}$, $\{\phi_i\}_{i=1}^N$, $\theta^{\rm R}$.
\end{algorithmic}
\end{algorithm}
\end{minipage}

\begin{center}
\begin{minipage}{0.8\textwidth}
\begin{algorithm}[H]
\caption{CORMA Deployment}
\label{alg:deployment}
\textbf{Input}: Initial constraint budget vector $\mathcal{C}_1 = (\kappa^1, \dots, \kappa^N)^T$, well-trained models $\theta^{\rm B}$, $\{\phi_i\}_{i=1}^N$, $\theta^{\rm R}$, deployment steps $T$. \\
\textbf{Initialize}: Initial state $s_1$, trajectory sequences $\{\tau^i=[\,]\}_{i=1}^N$, $\hat{\tau}=[\,]$.
\begin{algorithmic}[1]
\FOR{$t = 1, \dots, T$}
    \STATE Observe current state $s_t$ from the environment.
    \FOR{each constraint $i\in\{1,\dots,N\}$ }
    \STATE Append $s_t$, $C^i_t$ to trajectory sequence $\tau^i$.
    \STATE Formulate current condition sequence $\tau^i_{-K:t}$ for $i$-th expert using recent length-$K$ history trajectory sequence.
    \STATE Predict RTG distribution $p_{\phi_i}(\cdot|\tau_{-K:t}^i)$ based on each expert network.
    \ENDFOR
    \STATE Compute normalized remaining safety budgets vector $\bm{\bar{C}}_t$.
    \STATE Compute normalized weights vector $\bm \alpha_t$ based on the router.
    \STATE Generate constraint-aware RTG $\hat{R}_t$ by sampling from the fused PoE distribution:
    $\hat{R}_t \sim \frac{1}{Z_t} \prod_{i=1}^N p_{\phi_i}(\cdot | \tau_{-K:t}^i)^{\alpha_t^i}.$
    \STATE Append $s_t$, $\hat{R}_t$, and $\{C_t^i\}_{i=1}^N$ to trajectory sequence $\hat\tau$.
    \STATE Formulate current condition sequence $\hat{\tau}_{-K:t}$ using recent length-$K$ history trajectory sequence.
    \STATE Sample action $a_t$ from the conditional action policy $\pi_{\theta^{\rm B}}(\cdot|\hat{\tau}_{-K:t})$.
    \STATE Execute action $a_t$ in the environment, observe reward $r_t$ and costs $\{c^i_t\}_{i=1}^N$.
    \FOR{each constraint $i\in\{1,\dots,N\}$}
    \STATE Update remaining safety budget: $C^i_{t+1} = C^i_t - c^i_t$.
    \ENDFOR
    \STATE Append $R_t,a_t$ to trajectory sequence $\{\tau^i\}_{i=1}^N$.
    \STATE Append $a_t$ to trajectory sequence $\hat\tau$.
\ENDFOR
\end{algorithmic}
\end{algorithm}
\end{minipage}
\end{center}
\section*{E.Experimental Details}
\subsection{E.1 Experimental Tasks and Datasets}
\begin{figure*}[t]
    \centering
    \includegraphics[width=0.75\textwidth, keepaspectratio]{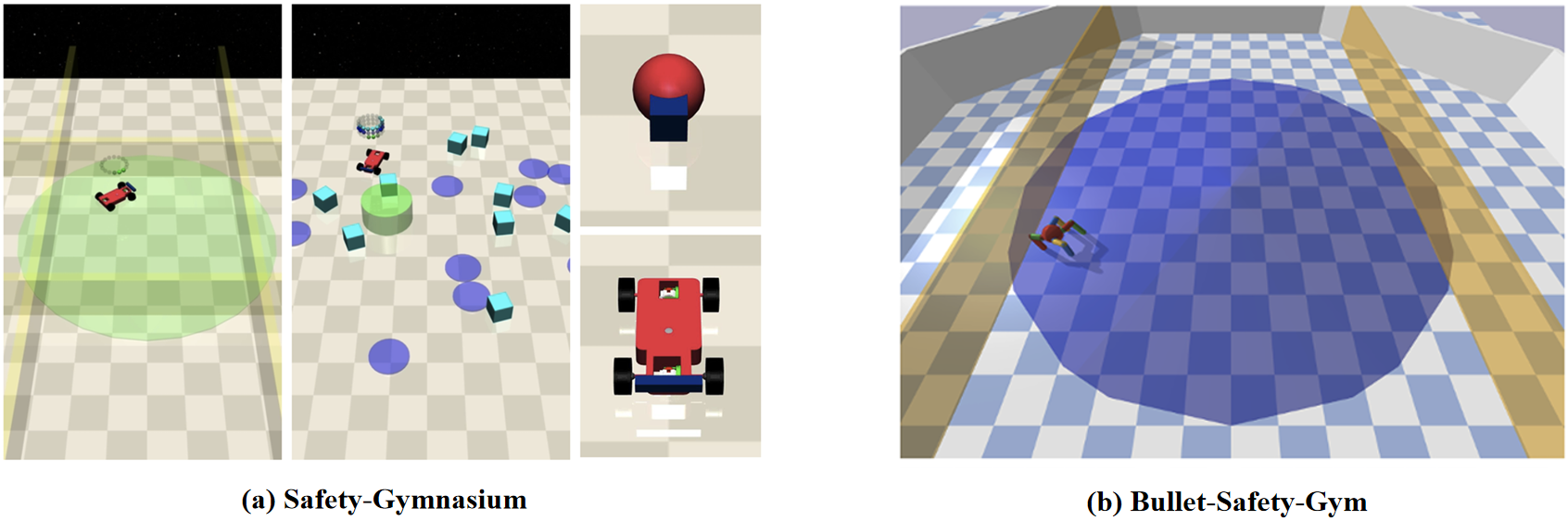}
    \caption{
    Visualization of the simulation environments and representative tasks (Liu et al., 2023a).
    }
    \label{fig:fine_tuning_convergence}
\end{figure*}

\textbf{Safety-Gymnasium} (Ji et al., 2023): Built on the MuJoCo physics simulator, Safety-Gymnasium provides safety-critical continuous-control environments with diverse robot-task combinations. In our experiments, we evaluate Point and Car agents on Button, Goal, and Push tasks, denoted as PB, CB, PG, CG, PP, and CP, respectively. In the Goal task, the agent is required to navigate toward a target location while avoiding unsafe regions. In the Push task, the agent must move an object to a designated goal position, which introduces more complex contact dynamics and longer-horizon planning. In the Button task, the agent needs to reach and activate target buttons scattered in the environment. These tasks involve heterogeneous safety requirements, including collision avoidance, high-velocity limitation, and low-velocity regulation, making them suitable for evaluating policy learning under multiple simultaneous constraints.

\textbf{Bullet-Safety-Gym} (Liu, 2023a): Built on the PyBullet physics simulator, Bullet-Safety-Gym contains lightweight yet challenging safety-control tasks for different robotic agents. We evaluate three Circle tasks with Ball, Car, and Drone agents, denoted as BC, CC, and DC, respectively. In the Circle task, the agent is encouraged to move along a circular trajectory while satisfying safety constraints related to boundary violation and velocity control. Compared with Safety-Gymnasium, Bullet-Safety-Gym provides shorter and more focused locomotion tasks, which allow us to examine whether an offline safe RL method can efficiently balance task reward and constraint satisfaction across different robot dynamics. For both benchmark suites, we construct two experimental settings: the -v2 setting includes boundary/collision and high-velocity constraints, while the more challenging -v3 setting further introduces a low-velocity constraint. Therefore, the evaluated tasks cover both two-constraint and three-constraint scenarios, enabling a comprehensive assessment of multi-constraint safety generalization.

The offline datasets used for each task are sourced from OSRL (Liu et al., 2023a). Fig. 1 visualizes some representative tasks of these
environments.
\subsection{E.2 Baselines}
We provide a more detailed introduction to the baselines used in our experiments in this section.
\begin{itemize}
    \item \textbf{BCQ-Lag} is a Lagrangian extension of Batch-Constrained Q-learning (BCQ) for offline safe reinforcement learning. BCQ constrains the learned policy to select actions close to the offline dataset distribution, thereby reducing extrapolation errors caused by out-of-distribution actions. In BCQ-Lag, safety constraints are handled by introducing Lagrangian multipliers into the policy optimization objective. For the multi-constraint setting, we extend the Lagrangian formulation with multiple dual variables, each corresponding to one safety constraint.

    \item \textbf{BEAR-Lag} is a Lagrangian variant of Bootstrapping Error Accumulation Reduction (BEAR). BEAR regularizes the learned policy by minimizing the divergence between policy actions and dataset actions, which helps reduce bootstrapping errors in offline reinforcement learning. BEAR-Lag further incorporates safety constraints through Lagrangian relaxation. Similar to BCQ-Lag, we adapt BEAR-Lag to multi-constraint scenarios by using separate Lagrangian multipliers for different constraints. 

    \item \textbf{PDCA} is a recent primal-dual critic-based offline constrained reinforcement learning method. It estimates the Lagrangian objective with learned reward and cost critics, and alternates between primal policy improvement and dual variable updates. Compared with classical Lagrangian baselines, PDCA provides a more principled primal-dual optimization framework for offline constrained policy learning. In our experiments, we extend PDCA to the multi-constraint setting by assigning an independent dual variable to each constraint. 

    \item \textbf{POCE} is a recent conservative offline constrained reinforcement learning algorithm designed for multi-constraint settings. It uses conservative value estimation to improve policy learning under safety constraints and reduces the risk of selecting unsafe out-of-distribution actions. Compared with BCQ-Lag and BEAR-Lag, POCE is more directly related to multi-constraint offline safe RL, since it explicitly considers multiple cost signals during policy optimization. 

    \item \textbf{OASIS} is a recent offline safe reinforcement learning method based on conditional distribution shaping. It aims to improve the quality of offline data by generating or reshaping trajectories toward safer and more rewarding regions. In our experiments, OASIS is adapted to multi-constraint tasks by treating a trajectory as feasible only when all constraint costs are below their corresponding safety thresholds. 

    \item \textbf{CAPS} is a constraint-adaptive policy switching method for offline safe reinforcement learning. It trains policies with different reward-cost trade-offs and chooses an appropriate policy according to the given safety budget during deployment. This makes CAPS suitable for evaluating constraint adaptation without retraining. Since the original formulation mainly considers scalar safety budgets, we adapt CAPS to the multi-constraint setting by selecting policies according to the maximum normalized constraint violation, i.e., Cost-N, and by requiring all normalized costs to satisfy the target thresholds.

    \item \textbf{CCAC} is a constraint-conditioned actor-critic method for offline safe reinforcement learning. It conditions the policy and value functions on safety thresholds, enabling the learned policy to adapt to different constraint budgets at test time. In our experiments, we extend CCAC to multi-constraint tasks by conditioning the policy on the multi-dimensional threshold vector and evaluating safety using the maximum normalized violation cost.

    \item \textbf{BC-Safe} is a behavior cloning baseline trained only on safe trajectories from the offline dataset. Specifically, trajectories whose costs satisfy all safety budgets are selected to train a supervised policy. This baseline provides a conservative reference for multi-constraint offline safe RL.

    \item \textbf{CDT-MC} is a multi-constraint adaptation of the Constrained Decision Transformer (CDT). It extends the standard Decision Transformer by incorporating cost-to-go (CTG) as an additional condition, allowing the policy to generate actions according to both the target return-to-go (RTG) and the remaining safety budget. In our multi-constraint experiments, CDT-MC is extended by concatenating multiple CTG tokens corresponding to different constraints. 

    \item \textbf{CoPDT-MC} is a multi-constraint adaptation of the recent sequence modeling method Constraint Prioritized Decision Transformer (CoPDT). CoPDT is designed to prioritize constraint satisfaction during offline policy learning. To handle our settings, CoPDT-MC extends this framework by incorporating multiple cost constraints into its sequence modeling architecture. 

    \item \textbf{CORDT} is a Transformer-based variant of our framework. It retains the constraint-specific expert routing and Product-of-Experts (PoE) fusion mechanism, but replaces the Mamba-based conditional action decoder with a Transformer backbone. This baseline is designed to isolate the contribution of the Mamba architecture. By comparing CORDT with CORMA, we can evaluate whether the linear-complexity selective state-space model improves sequence modeling efficiency and long-horizon decision-making under multi-constraint conditions.

    \item \textbf{CORMA w/o PoE} is an ablation baseline that completely removes the constraint-specific experts and the PoE fusion mechanism. It directly utilizes the empirical RTG and concatenated CTGs as inputs for the Mamba action decoder, without explicitly modeling the feasible RTG distribution. This baseline is used to evaluate the necessity and effectiveness of our constraint-routed RTG estimation module.

    \item \textbf{CORMA-S} is an ablation baseline that removes the constraint-decoupled architecture. Instead of using multiple constraint-specific experts, it employs a single monolithic expert to directly model the joint RTG distribution conditioned on all CTG tokens simultaneously. This helps verify whether a single unified network is sufficient to disentangle conflicting safety requirements compared to our decoupled routing mechanism.
\end{itemize}
\subsection{E.3 Safety-Budget Shift Regions Settings}

As introduced in the main text, to evaluate zero-shot
generalization under unseen safety-budget combinations, we
partition the safety-budget space according to the minimum
normalized relative distance from an evaluation budget vector
$\boldsymbol{\kappa}$ to the seen budget region
$\mathcal{K}_{\mathrm{seen}}$:
\begin{equation}
\label{eq:budget_distance}
d(\boldsymbol{\kappa},\mathcal{K}_{\mathrm{seen}})
=
\min_{\boldsymbol{\kappa}^{\prime}
\in\mathcal{K}_{\mathrm{seen}}}
\sqrt{
\frac{1}{N}
\sum_{i=1}^{N}
\left(
\frac{
\kappa^i-\kappa^{\prime i}
}{
\kappa^{\prime i}
}
\right)^2
}.
\end{equation}
Here, all safety-budget components are assumed to be positive.
Based on this distance metric, we divide the evaluation space
into four continuous regions with progressively increasing
deviations: seen, slight, moderate, and severe. The region
boundaries are predefined as
$d_1=0.05$, $d_2=0.10$, and $d_3=0.15$.
The four regions are defined as: (1)Seen region: $d(\boldsymbol{\kappa},\mathcal{K}_{\mathrm{seen}})=0$; (2)Slight-shift region: $d(\boldsymbol{\kappa},\mathcal{K}_{\mathrm{seen}})\in(0,d_1]$; (3)Moderate-shift region: $d(\boldsymbol{\kappa},\mathcal{K}_{\mathrm{seen}})\in(d_1,d_2]$; (4)Severe-shift region: $d(\boldsymbol{\kappa},\mathcal{K}_{\mathrm{seen}})\in(d_2,d_3]$.
We generate
diverse budget combinations by perturbing a reference boundary
budget
$\boldsymbol{\kappa}^{\mathrm{ref}}$
along multiple directions. For an evaluation region
$(d_{\mathrm{low}},d_{\mathrm{high}}]$, we sample a target
perturbation magnitude
$\delta\sim\mathcal{U}
(d_{\mathrm{low}},d_{\mathrm{high}})$.
A candidate evaluation budget is then constructed using a
constraint-specific perturbation vector
$\boldsymbol{\Delta}
=[\Delta^1,\ldots,\Delta^N]^\top.$
The perturbation vector is normalized according to
\begin{equation}
\label{eq:perturbation_norm}
\sqrt{
\frac{1}{N}
\sum_{i=1}^{N}
(\Delta^i)^2
}
=
\delta.
\end{equation}
Based on the sampled perturbation magnitude $\delta$, we consider
three perturbation scenarios to simulate heterogeneous constraint
variations:
\begin{enumerate}
    \item \textbf{Uniform Shift:}
    All constraint budgets are changed by the same proportion and
    in the same direction. They are either jointly tightened or
    jointly relaxed.

    \item \textbf{Asymmetric Shift:}
    One constraint budget is substantially tightened or relaxed,
    while the remaining constraints stay at the seen boundary.

    \item \textbf{Conflicting Shift:}
    One or more constraint budgets are tightened, while the
    remaining budgets are relaxed.
\end{enumerate}

\paragraph{-v2 tasks.}
The -v2 tasks contain the collision/boundary constraint and the
high-velocity constraint. The reference budget is represented as
$\boldsymbol{\kappa}_{\mathrm{v2}}^{\mathrm{ref}}
=
\left[
\kappa_{\mathrm{ref}}^{\mathrm{col}},
\kappa_{\mathrm{ref}}^{\mathrm{high}}
\right]^\top.$
For a sampled perturbation magnitude $\delta$, representative
candidate budgets are constructed as
\begin{align}
\boldsymbol{\kappa}_{\mathrm{v2}}^{\mathrm{uni}}
&=
\left[
(1\pm\Delta^{\mathrm{col}})\kappa_{\mathrm{ref}}^{\mathrm{col}},
(1\pm\Delta^{\mathrm{high}})\kappa_{\mathrm{ref}}^{\mathrm{high}}
\right]^\top,
\\
\boldsymbol{\kappa}_{\mathrm{v2}}^{\mathrm{asy}}
&=
\left[
(1\pm\sqrt{2}\Delta^{\mathrm{col}})
\kappa_{\mathrm{ref}}^{\mathrm{col}},
\kappa_{\mathrm{ref}}^{\mathrm{high}}
\right]^\top,
\\
\boldsymbol{\kappa}_{\mathrm{v2}}^{\mathrm{conf}}
&=
\left[
(1\pm\Delta^{\mathrm{col}})
\kappa_{\mathrm{ref}}^{\mathrm{col}},
(1\mp\Delta^{\mathrm{high}})
\kappa_{\mathrm{ref}}^{\mathrm{high}}
\right]^\top.
\end{align}

\paragraph{-v3 tasks.}
The -v3 tasks additionally include the low-velocity constraint.
The corresponding reference budget is
\[
\boldsymbol{\kappa}_{\mathrm{v3}}^{\mathrm{ref}}
=
\left[
\kappa_{\mathrm{ref}}^{\mathrm{col}},
\kappa_{\mathrm{ref}}^{\mathrm{high}},
\kappa_{\mathrm{ref}}^{\mathrm{low}}
\right]^\top.
\]
Representative candidate budgets are constructed as
\begin{align}
\boldsymbol{\kappa}_{\mathrm{v3}}^{\mathrm{uni}}
&=
\left[
(1\pm\Delta^{\mathrm{col}})
\kappa_{\mathrm{ref}}^{\mathrm{col}},
(1\pm\Delta^{\mathrm{high}})
\kappa_{\mathrm{ref}}^{\mathrm{high}},
(1\pm\Delta^{\mathrm{low}})
\kappa_{\mathrm{ref}}^{\mathrm{low}}
\right]^\top,
\\
\boldsymbol{\kappa}_{\mathrm{v3}}^{\mathrm{asy}}
&=
\left[
(1\pm\sqrt{3}\Delta^{\mathrm{col}})
\kappa_{\mathrm{ref}}^{\mathrm{col}},
\kappa_{\mathrm{ref}}^{\mathrm{high}},
\kappa_{\mathrm{ref}}^{\mathrm{low}}
\right]^\top,
\\
\boldsymbol{\kappa}_{\mathrm{v3}}^{\mathrm{conf}}
&=
\left[
(1\pm\Delta^{\mathrm{col}})
\kappa_{\mathrm{ref}}^{\mathrm{col}},
(1\pm\Delta^{\mathrm{high}})
\kappa_{\mathrm{ref}}^{\mathrm{high}},
(1\mp\Delta^{\mathrm{low}})
\kappa_{\mathrm{ref}}^{\mathrm{low}}
\right]^\top.
\end{align}

\subsection{E.4 Hyperparameters}
The training and deployment of CORMA both involve the selection of hyperparameters. To ensure reproducibility, this section outlines the specific hyperparameters used in our experiments. CORMA is implemented within the OSRL framework, and the default parameters are retained for any hyperparameters not explicitly mentioned in Table \ref{tab:hyperparameters}.
\begin{table}[htbp]
\centering
\begin{tabular}{lll}
\toprule
 & Hyperparameter & Value \\
\midrule
\multirow{7}{*}{conditional action decoder $\theta^B$} 
 & Mamba num layers & 2 \\
 & Mamba embedding dim & 128 \\
 & Mamba expand factor & 2 \\
 & Mamba convolution size & 4 \\
 & update steps & 30000 \\
 & batch size & 1024 \\
 & learning rate & 0.0001 \\
\midrule
\multirow{7}{*}{constraint-specific RTG experts $\phi_i$} 
 & Mamba num layers & 2 \\
 & Mamba embedding dim & 128 \\
 & Mamba expand factor & 2 \\
 & Mamba convolution size & 4 \\
 & update steps & 60000 \\
 & batch size & 1024 \\
 & learning rate & 0.0001 \\
\midrule
\multirow{4}{*}{gating router network $\theta^R$} 
 & router hidden layers & [256, 256, 256] \\
 & update steps & 20000 \\
 & batch size & 1024 \\
 & learning rate & 0.0001 \\
\midrule
\multirow{5}{*}{policy learning and deployment} 
 & sequence len $K$ & 50 \\
 & $C_{\rho}^i$ & 0.5 \\
 & $R_{\eta}^i$ & 0.8 \\
 & $\lambda$ & 0.001 \\
 & Dropout & 0.1 \\
 & Weight decay & 0.01 \\
\bottomrule
\end{tabular}
\caption{Hyperparameter choices of CORMA.}
\label{tab:hyperparameters}
\end{table}
\section*{F.More Experimental Results}

\subsection*{F.1 Analysis on Safe Policy Learning under Multiple Constraints}

\begin{table}[!t]
\centering
\renewcommand{\arraystretch}{1.1}
\setlength{\tabcolsep}{1pt}

\begin{tabular}{llcccccccccc}
\toprule
\multirow{3}{*}{\textbf{Task}} & \multirow{3}{*}{\textbf{Method}} & \multicolumn{10}{c}{\textbf{Safety-budget Combinations} } \\
\cmidrule{3-12}
& & \multicolumn{2}{c}{\textbf{S1}}
& \multicolumn{2}{c}{\textbf{S2}}
& \multicolumn{2}{c}{\textbf{S3}}
& \multicolumn{2}{c}{\textbf{S4}}
& \multicolumn{2}{c}{\textbf{S5}} \\
\cmidrule{3-12}
& & Reward & Cost-$N$ & Reward & Cost-$N$ & Reward & Cost-$N$ & Reward & Cost-$N$ & Reward & Cost-$N$ \\
\midrule

\multirow{14}{*}{\textbf{BC-v2}}
& BCQ-Lag
& 88.6 & 1.06 & 82.4 & 1.21 & 85.7 & 1.48
& \textbf{80.1} & \textbf{0.94} & 81.2 & 1.03 \\

& BEAR-Lag
& 84.7 & 1.12 & 88.2 & 1.07 & 91.5 & 1.01
& 86.4 & 1.09 & 87.1 & 1.15 \\

& PDCA
& 88.9 & 1.15 & \textbf{83.8} & \textbf{0.96}
& \underline{\textbf{88.5}} & \underline{\textbf{0.99}}
& 81.2 & 1.05 & \textbf{82.6} & \textbf{0.97} \\

& POCE
& 89.6 & 1.02 & \textbf{84.5} & \textbf{0.99}
& 89.2 & 1.16 & 82.2 & 1.01 & 87.0 & 1.19 \\

& OASIS
& \textbf{85.8} & \textbf{0.99}
& \textbf{80.6} & \textbf{0.98}
& \textbf{85.1} & \textbf{0.93}
& \underline{\textbf{88.2}} & \underline{\textbf{0.98}}
& \underline{\textbf{89.5}} & \underline{\textbf{0.97}} \\

& CAPS
& \textbf{73.5} & \textbf{0.94}
& \textbf{78.0} & \textbf{0.92}
& \textbf{82.8} & \textbf{0.98}
& \textbf{75.4} & \textbf{0.93}
& \textbf{76.8} & \textbf{0.95} \\

& CCAC
& \textbf{80.2} & \textbf{0.96}
& \textbf{85.6} & \textbf{0.94}
& \textbf{80.4} & \textbf{0.91}
& \textbf{82.8} & \textbf{0.95}
& \textbf{84.1} & \textbf{0.97} \\

& BC-Safe
& 71.8 & 1.79
& \textbf{76.6} & \textbf{0.99}
& 77.3 & 1.54
& 73.5 & 1.37
& \textbf{76.1} & \textbf{0.98} \\

& CDT-MC
& 70.9 & 1.58
& \textbf{75.4} & \textbf{0.98}
& 76.2 & 1.43
& 73.2 & 1.54
& 74.6 & 1.62 \\

& CoPDT-MC
& 91.5 & 1.18
& 86.8 & 1.12
& 81.4 & 1.08
& 83.9 & 1.15
& 85.0 & 1.20 \\

& CORDT
& \textbf{88.2} & \textbf{0.96}
& 83.5 & 1.15
& 87.9 & 1.48
& \textbf{80.5} & \textbf{0.99}
& 82.4 & 1.65 \\

& CORMA w/o PoE
& \textbf{70.4} & \textbf{0.98}
& 74.7 & 1.22
& 78.3 & 1.28
& 72.8 & 1.25
& \textbf{73.6} & \textbf{0.99} \\

& CORMA-S
& 87.5 & 1.28
& 82.8 & 1.03
& \textbf{87.2} & \textbf{0.98}
& 89.6 & 1.06
& 81.5 & 1.12 \\

& \textbf{CORMA (Ours)}
& \underline{\textbf{89.8}} & \underline{\textbf{0.95}}
& \underline{\textbf{88.6}} & \underline{\textbf{0.93}}
& \textbf{80.1} & \textbf{0.87}
& \textbf{85.2} & \textbf{0.96}
& \textbf{86.5} & \textbf{0.98} \\
\midrule

\multirow{14}{*}{\textbf{CC-v2}}
& BCQ-Lag
& 72.8 & 1.09
& \textbf{76.4} & \textbf{0.99}
& \textbf{78.1} & \textbf{0.96}
& 73.7 & 1.04
& 74.5 & 1.01 \\

& BEAR-Lag
& 82.6 & 1.04
& 84.9 & 1.08
& 86.1 & 1.03
& 81.7 & 1.06
& 83.4 & 1.05 \\

& PDCA
& 79.4 & 1.11
& \textbf{82.1} & \textbf{0.96}
& \textbf{83.7} & \textbf{0.95}
& \textbf{80.6} & \textbf{0.98}
& \underline{\textbf{80.9}} & \underline{\textbf{0.97}} \\

& POCE
& 80.3 & 1.27
& 82.8 & 1.05
& 84.4 & 1.09
& 81.2 & 1.08
& 78.9 & 1.11 \\

& OASIS
& \textbf{76.8} & \textbf{0.97}
& \textbf{79.2} & \textbf{0.99}
& 80.5 & 1.05
& \textbf{77.4} & \textbf{0.98}
& 85.1 & 1.07 \\

& CAPS
& \textbf{74.6} & \textbf{0.96}
& \textbf{76.8} & \textbf{0.98}
& \textbf{77.9} & \textbf{0.97}
& \textbf{74.9} & \textbf{1.00}
& \textbf{73.3} & \textbf{0.99} \\

& CCAC
& \textbf{80.8} & \textbf{0.95}
& \underline{\textbf{82.7}} & \underline{\textbf{0.94}}
& \textbf{84.2} & \textbf{0.95}
& \underline{\textbf{83.1}} & \underline{\textbf{0.93}}
& \textbf{79.6} & \textbf{0.97} \\

& BC-Safe
& 74.1 & 1.07
& 79.6 & 1.44
& 81.3 & 1.29
& 77.0 & 1.31
& 73.9 & 1.36 \\

& CDT-MC
& 71.9 & 1.09
& 80.7 & 1.24
& 79.8 & 1.39
& 74.3 & 1.28
& 72.8 & 1.18 \\

& CoPDT-MC
& \textbf{76.5} & \textbf{0.97}
& 80.1 & 1.18
& \textbf{82.3} & \textbf{0.95}
& 77.6 & 1.22
& \textbf{75.4} & \textbf{0.99} \\

& CORDT
& 81.1 & 1.12
& 84.5 & 1.19
& \underline{\textbf{86.7}} & \underline{\textbf{0.99}}
& 82.0 & 1.01
& 78.8 & 1.13 \\

& CORMA w/o PoE
& \textbf{79.9} & \textbf{0.97}
& \textbf{81.6} & \textbf{0.99}
& 83.8 & 1.16
& 79.4 & 1.02
& \textbf{78.2} & \textbf{0.98} \\

& CORMA-S
& 76.1 & 1.17
& 80.4 & 1.25
& 82.0 & 1.30
& 77.9 & 1.18
& 75.5 & 1.25 \\

& \textbf{CORMA (Ours)}
& \underline{\textbf{81.6}} & \underline{\textbf{0.97}}
& \textbf{81.9} & \textbf{0.95}
& \textbf{85.1} & \textbf{0.92}
& \textbf{82.6} & \textbf{0.97}
& \textbf{80.4} & \textbf{0.96} \\
\midrule

\multirow{14}{*}{\textbf{DC-v2}}
& BCQ-Lag
& 77.3 & 1.14
& \textbf{71.5} & \textbf{0.98}
& \textbf{75.4} & \textbf{0.98}
& \textbf{79.1} & \textbf{0.97}
& \textbf{70.2} & \textbf{0.97} \\

& BEAR-Lag
& 79.4 & 1.03
& 74.8 & 1.07
& \textbf{79.6} & \textbf{0.99}
& 71.9 & 1.29
& 73.2 & 1.04 \\

& PDCA
& 70.2 & 1.05
& \textbf{76.6} & \textbf{0.99}
& \textbf{82.0} & \textbf{0.94}
& 73.1 & 1.02
& 74.5 & 1.08 \\

& POCE
& 71.0 & 1.01
& 77.5 & 1.06
& \textbf{83.2} & \textbf{0.91}
& 74.2 & 1.19
& 75.6 & 1.03 \\

& OASIS
& \underline{\textbf{78.0}} & \underline{\textbf{0.98}}
& \textbf{74.1} & \textbf{0.94}
& \textbf{80.1} & \textbf{0.89}
& \textbf{71.0} & \textbf{0.96}
& \textbf{72.2} & \textbf{0.99} \\

& CAPS
& \textbf{76.3} & \textbf{0.93}
& \textbf{72.2} & \textbf{0.95}
& \textbf{78.5} & \textbf{0.85}
& \textbf{79.8} & \textbf{0.92}
& \textbf{80.1} & \textbf{0.94} \\

& CCAC
& \textbf{72.3} & \textbf{0.96}
& \textbf{78.6} & \textbf{0.94}
& \underline{\textbf{84.5}} & \underline{\textbf{0.88}}
& \textbf{75.4} & \textbf{0.95}
& \textbf{76.3} & \textbf{0.97} \\

& BC-Safe
& 75.6 & 1.03
& 79.2 & 1.35
& \textbf{72.8} & \textbf{0.97}
& 77.5 & 1.27
& \textbf{78.4} & \textbf{0.97} \\

& CDT-MC
& 83.2 & 1.05
& \textbf{79.6} & \textbf{0.96}
& \textbf{75.7} & \textbf{0.98}
& 76.8 & 1.01
& \textbf{78.3} & \textbf{0.99} \\

& CoPDT-MC
& 74.3 & 1.25
& 80.5 & 1.20
& 86.0 & 1.14
& 77.2 & 1.23
& 78.8 & 1.30 \\

& CORDT
& \textbf{71.5} & \textbf{0.96}
& 77.2 & 1.28
& \textbf{82.6} & \textbf{0.97}
& \textbf{74.2} & \textbf{0.96}
& 75.9 & 1.10 \\

& CORMA w/o PoE
& \textbf{72.8} & \textbf{0.92}
& \textbf{77.6} & \textbf{0.96}
& 71.9 & 1.16
& \textbf{75.4} & \textbf{0.99}
& \textbf{76.6} & \textbf{0.95} \\

& CORMA-S
& 78.8 & 1.06
& 75.1 & 1.01
& \textbf{81.4} & \textbf{0.96}
& 71.6 & 1.04
& 73.0 & 1.17 \\

& \textbf{CORMA (Ours)}
& \textbf{71.6} & \textbf{0.96}
& \underline{\textbf{79.7}} & \underline{\textbf{0.93}}
& \textbf{83.6} & \textbf{0.95}
& \underline{\textbf{82.1}} & \underline{\textbf{0.97}}
& \underline{\textbf{83.9}} & \underline{\textbf{0.96}} \\
\bottomrule
\end{tabular}
\caption{Detailed evaluation under seen two-safety-budget combinations. Reward is normalized to $0$--$100$, and the safety threshold is normalized to 1. S1--S5 denote threshold combinations appearing in the offline dataset. \textbf{Bold} indicates constraint satisfaction ($c\leq1$), and \underline{\textbf{underlined bold}} indicates the highest reward among feasible methods.}
\label{tab:seen_threshold_v2_perfect}
\end{table}

\begin{table*}[!t]
\centering
\renewcommand{\arraystretch}{1.1}
\setlength{\tabcolsep}{1pt}

\begin{tabular}{llcccccccccc}
\toprule
\multirow{3}{*}{\textbf{Task}} & \multirow{3}{*}{\textbf{Method}} & \multicolumn{10}{c}{\textbf{safety-budget combinations} } \\
\cmidrule{3-12}
& & \multicolumn{2}{c}{\textbf{S1}}
& \multicolumn{2}{c}{\textbf{S2}}
& \multicolumn{2}{c}{\textbf{S3}}
& \multicolumn{2}{c}{\textbf{S4}}
& \multicolumn{2}{c}{\textbf{S5}} \\
\cmidrule{3-12}
& & Reward & Cost-$N$ & Reward & Cost-$N$ & Reward & Cost-$N$ & Reward & Cost-$N$ & Reward & Cost-$N$ \\
\midrule

\multirow{14}{*}{\textbf{BC-v3}}
& BCQ-Lag
& 72.6 & 1.82
& 77.8 & 1.63
& 75.1 & 1.48
& 73.4 & 1.91
& 73.6 & 1.71 \\

& BEAR-Lag
& 84.7 & 1.61
& 88.9 & 1.86
& 86.1 & 1.72
& 83.8 & 2.24
& 88.0 & 2.02 \\

& PDCA
& 76.8 & 1.48
& 81.5 & 1.60
& 80.3 & 1.49
& 77.1 & 1.94
& 77.8 & 1.88 \\

& POCE
& 77.2 & 1.45
& 82.6 & 1.29
& 80.4 & 1.18
& 78.5 & 1.50
& 78.3 & 1.42 \\

& OASIS
& 71.9 & 1.24
& 76.4 & 1.12
& \textbf{74.8} & \textbf{0.99}
& 72.3 & 1.32
& 73.6 & 1.24 \\

& CAPS
& 69.7 & 1.17
& 73.4 & 1.08
& \textbf{72.8} & \textbf{0.97}
& 70.2 & 1.25
& 70.4 & 1.22 \\

& CCAC
& \underline{\textbf{79.4}}
& \underline{\textbf{0.97}}
& 75.8 & 1.16
& \textbf{82.0} & \textbf{0.99}
& 74.6 & 1.13
& 75.7 & 1.15 \\

& BC-Safe
& 80.6 & 1.21
& 84.9 & 1.13
& \underline{\textbf{83.2}}
& \underline{\textbf{0.96}}
& 81.1 & 1.19
& 82.2 & 1.26 \\

& CDT-MC
& 75.4 & 1.73
& 80.8 & 1.55
& 82.1 & 1.38
& 77.6 & 1.69
& 75.1 & 1.75 \\

& CoPDT-MC
& 77.9 & 1.58
& 82.7 & 1.43
& 81.4 & 1.31
& 78.2 & 1.56
& 78.8 & 1.57 \\

& CORDT
& 78.6 & 1.36
& 83.1 & 1.22
& \textbf{81.7} & \textbf{0.98}
& 79.4 & 1.41
& 78.2 & 1.43 \\

& CORMA w/o PoE
& 73.5 & 1.57
& 78.9 & 1.39
& 77.1 & 1.24
& 74.6 & 1.68
& 74.9 & 1.59 \\

& CORMA-S
& 71.6 & 1.98
& 76.9 & 1.74
& 74.4 & 1.59
& 72.5 & 2.04
& 73.6 & 2.00 \\

& \textbf{CORMA (Ours)}
& \textbf{78.7} & \textbf{0.98}
& \underline{\textbf{83.0}}
& \underline{\textbf{0.99}}
& \textbf{79.8} & \textbf{0.97}
& 82.5 & 1.01
& \underline{\textbf{80.5}}
& \underline{\textbf{0.93}} \\
\midrule

\multirow{14}{*}{\textbf{CC-v3}}
& BCQ-Lag
& 64.9 & 1.97
& 67.4 & 1.77
& 74.0 & 1.91
& 69.6 & 1.85
& 72.5 & 1.83 \\

& BEAR-Lag
& 73.5 & 2.01
& 67.0 & 1.62
& 77.9 & 1.79
& 70.4 & 1.34
& 77.2 & 1.44 \\

& PDCA
& 61.1 & 1.69
& 73.3 & 1.90
& 76.1 & 1.87
& 74.5 & 1.77
& 73.0 & 1.72 \\

& POCE
& 73.3 & 1.30
& 69.8 & 1.43
& 79.6 & 1.54
& 66.0 & 1.42
& 73.3 & 1.24 \\

& OASIS
& 70.4 & 1.13
& 62.4 & 1.22
& 73.2 & 1.17
& 76.8 & 1.56
& 74.7 & 1.12 \\

& CAPS
& \textbf{63.7} & \textbf{0.94}
& \textbf{68.1} & \textbf{0.99}
& \textbf{66.4} & \textbf{0.98}
& 64.9 & 1.08
& 65.9 & 1.19 \\

& CCAC
& 68.2 & 1.65
& 74.5 & 1.26
& 76.1 & 1.44
& 72.7 & 1.19
& 64.5 & 1.16 \\

& BC-Safe
& 68.3 & 1.69
& 70.6 & 1.61
& 72.9 & 1.70
& 75.2 & 1.71
& 76.0 & 1.58 \\

& CDT-MC
& 69.4 & 1.12
& \underline{\textbf{74.1}}
& \underline{\textbf{0.97}}
& 77.3 & 1.08
& 70.6 & 1.14
& 67.6 & 1.09 \\

& CoPDT-MC
& 73.8 & 1.24
& 69.5 & 1.38
& 70.8 & 1.24
& 68.3 & 1.52
& 74.1 & 1.53 \\

& CORDT
& 75.0 & 1.05
& \textbf{73.8} & \textbf{0.89}
& \underline{\textbf{74.5}}
& \underline{\textbf{0.94}}
& \underline{\textbf{71.2}}
& \underline{\textbf{0.92}}
& 67.5 & 1.10 \\

& CORMA w/o PoE
& 73.9 & 1.06
& \textbf{67.8} & \textbf{0.94}
& \textbf{72.6} & \textbf{0.98}
& 69.7 & 1.07
& 71.5 & 1.15 \\

& CORMA-S
& 75.2 & 1.67
& 65.6 & 1.73
& 69.9 & 1.46
& 77.7 & 1.94
& 71.6 & 1.57 \\

& \textbf{CORMA (Ours)}
& \underline{\textbf{72.4}}
& \underline{\textbf{0.95}}
& 68.1 & 1.09
& \textbf{70.3} & \textbf{0.96}
& 69.7 & 1.09
& \underline{\textbf{73.5}}
& \underline{\textbf{0.93}} \\
\midrule

\multirow{14}{*}{\textbf{DC-v3}}
& BCQ-Lag
& 93.3 & 1.77
& 90.1 & 2.03
& 94.4 & 1.55
& 94.8 & 1.73
& 87.9 & 1.85 \\

& BEAR-Lag
& 95.5 & 2.63
& 92.8 & 2.85
& 96.0 & 2.61
& 93.9 & 2.94
& 94.3 & 2.78 \\

& PDCA
& 88.8 & 1.46
& 87.6 & 1.22
& 92.2 & 1.59
& 87.5 & 1.79
& 88.2 & 1.71 \\

& POCE
& 91.4 & 1.53
& 85.3 & 1.31
& 85.1 & 1.18
& 84.4 & 1.28
& 87.8 & 1.45 \\

& OASIS
& 82.9 & 1.75
& 82.4 & 1.89
& 79.5 & 1.71
& 81.7 & 1.83
& 81.4 & 1.72 \\

& CAPS
& 79.3 & 1.06
& 80.2 & 1.13
& 72.1 & 1.44
& 78.9 & 1.22
& 77.5 & 1.35 \\

& CCAC
& 80.4 & 1.17
& \textbf{83.7} & \textbf{0.96}
& \textbf{84.9} & \textbf{0.97}
& \textbf{85.9} & \textbf{0.99}
& 86.2 & 1.04 \\

& BC-Safe
& 41.2 & 1.05
& \textbf{61.6} & \textbf{0.91}
& \textbf{52.1} & \textbf{0.98}
& 45.5 & 1.04
& \textbf{54.6} & \textbf{0.93} \\

& CDT-MC
& 88.7 & 1.40
& 85.8 & 1.11
& 82.4 & 1.32
& 86.3 & 1.50
& 83.4 & 1.41 \\

& CoPDT-MC
& 86.9 & 1.44
& 85.0 & 1.47
& 83.3 & 1.52
& 85.1 & 1.78
& 81.2 & 1.54 \\

& CORDT
& \textbf{82.4} & \textbf{0.99}
& \underline{\textbf{86.4}}
& \underline{\textbf{0.90}}
& \textbf{82.5} & \textbf{0.98}
& 85.1 & 1.12
& \underline{\textbf{81.1}}
& \underline{\textbf{0.83}} \\

& CORMA w/o PoE
& \textbf{84.1} & \textbf{0.99}
& 83.6 & 1.08
& 81.3 & 1.06
& 79.6 & 1.25
& 83.2 & 1.03 \\

& CORMA-S
& 76.9 & 1.50
& 79.8 & 1.92
& 80.9 & 1.22
& 82.6 & 1.62
& 81.4 & 1.27 \\

& \textbf{CORMA (Ours)}
& \underline{\textbf{88.4}}
& \underline{\textbf{0.93}}
& \textbf{86.1} & \textbf{0.94}
& \underline{\textbf{90.9}}
& \underline{\textbf{0.97}}
& \underline{\textbf{88.8}}
& \underline{\textbf{0.95}}
& 85.7 & 1.02 \\
\bottomrule
\end{tabular}

\caption{Detailed evaluation under seen three-safety-budget combinations. Reward is normalized to $0$--$100$, and the safety threshold is normalized to 1. S1--S5 denote threshold combinations appearing in the offline dataset. \textbf{Bold} indicates constraint satisfaction ($c\leq1$), and \underline{\textbf{underlined bold}} indicates the highest reward among feasible methods.} 
\label{tab:seen_threshold_v3_perfect}
\end{table*}

Tables~\ref{tab:seen_threshold_v2_perfect} and
\ref{tab:seen_threshold_v3_perfect} report the performance of
different algorithms under the seen safety-budget combinations
contained in the offline datasets. The v2 tasks involve two
constraint dimensions, whereas the v3 tasks additionally introduce
a low-velocity constraint, resulting in more complex interactions
among heterogeneous safety requirements. These experiments evaluate
whether each method can adapt to different multi-dimensional safety
budgets observed during training while maintaining both constraint
satisfaction and competitive task performance.
Overall, CORMA achieves the most stable reward--safety trade-off
across the evaluated settings. Among the 30 task--budget
combinations, CORMA satisfies the normalized safety threshold in
26 cases, substantially more than the other sequence-modeling
variants. Averaged over all combinations, CORMA achieves a
normalized reward of approximately $81.4$ with an average Cost-$N$
of $0.96$. These results indicate that CORMA can adapt its behavior
to different safety-budget configurations without relying on a
single fixed constraint setting.

On the v2 tasks, CORMA maintains Cost-$N$ below $1.0$ under all
15 safety-budget combinations. It also achieves the best
safety-performance trade-off in six combinations, including S1 and S2 on BC-v2, S1 on CC-v2, and S2, S4, and S5 on DC-v2. Although several conventional offline safe RL methods, such
as OASIS, CAPS, and CCAC, also obtain feasible results under some
two-constraint settings, their reward--safety trade-offs vary across
tasks and budget combinations. In contrast, CDT-MC and CoPDT-MC
frequently violate the safety threshold despite attaining relatively
high rewards, indicating that directly conditioning a sequence model
on multiple CTG tokens is insufficient for reliably coordinating
heterogeneous constraints.

The three-constraint results further reveal the increased difficulty
of multi-constraint policy learning. CORMA remains feasible in 11
of the 15 v3 task--budget combinations and achieves an average
Cost-$N$ of approximately $0.98$. On BC-v3 and DC-v3, CORMA
satisfies the constraints in four of the five combinations, while
the remaining violations are marginal, with Cost-$N$ values of
$1.01$ and $1.02$, respectively. CC-v3 is more challenging:
CORMA remains feasible under S1, S3, and S5, but exhibits violations
under S2 and S4. Nevertheless, its Cost-$N$ remains close to the
safety boundary in most cases, while many competing methods incur
considerably larger violations. In particular, CDT-MC is feasible
in only one v3 combination, CoPDT-MC fails to satisfy the threshold
in all v3 combinations, and CORMA-S remains unsafe throughout the
three-constraint evaluations.

The comparisons among the CORMA variants further clarify the
contribution of the proposed architecture. Removing the
constraint-routed PoE module reduces the number of feasible
combinations and increases the average Cost-$N$, especially on the
v3 tasks. Similarly, CORMA-S, which replaces the constraint-specific
experts with a single shared RTG predictor, exhibits severe
constraint violations, suggesting that a unified expert cannot
adequately represent the heterogeneous and potentially conflicting
effects of multiple constraints. These results support the use of
constraint-wise expert specialization and adaptive PoE fusion for
estimating a feasible RTG under different safety-budget
combinations.
Compared with CORDT, CORMA also exhibits more robust safety
performance across the evaluated budgets. CORDT satisfies the
constraint threshold in 14 of the 30 combinations, whereas CORMA
is feasible in 26 combinations. Moreover, CORDT obtains an average
Cost-$N$ of approximately $1.10$, compared with $0.96$ for CORMA.
Because both methods use the same constraint-routed PoE mechanism,
this improvement indicates that the Mamba-based action decoder
provides more stable long-horizon action generation under varying
multi-constraint conditions. Nevertheless, the remaining violations
of CORMA on several difficult v3 combinations also show that
simultaneously satisfying highly restrictive and heterogeneous
budgets remains challenging.

These results demonstrate that CORMA provides a more reliable
mechanism for learning budget-conditioned policies from
multi-constraint offline datasets. Its advantage is reflected not
only in high rewards, but more importantly in its substantially
higher feasibility rate and lower average Cost-$N$ across diverse
seen safety-budget combinations. The improvements over CORMA w/o
PoE, CORMA-S, and CORDT jointly verify the complementary benefits
of constraint-specific RTG experts, router-guided PoE fusion, and
the Mamba-based action decoder.

\subsection{F.2 Analysis on Trajectory Stitching Ability}
\begin{table}[!t]
\centering
\renewcommand{\arraystretch}{1.1}
\setlength{\tabcolsep}{2.5pt} 
\begin{tabular}{lcccccccc}
\toprule
\multirow{2}{*}{\textbf{Task}}
& \multicolumn{2}{c}{\textbf{CDT-MC}}
& \multicolumn{2}{c}{\textbf{CORMA w/o PoE}}
& \multicolumn{2}{c}{\textbf{CORDT}}
& \multicolumn{2}{c}{\textbf{CORMA (Ours)}} \\
\cmidrule(lr){2-3} \cmidrule(lr){4-5} \cmidrule(lr){6-7} \cmidrule(lr){8-9}
& Reward & Cost-$N$
& Reward & Cost-$N$
& Reward & Cost-$N$
& Reward & Cost-$N$ \\  
\midrule

BC-v2
& \textbf{71.2} & \textbf{0.98}
& 72.8 & 1.22
& \textbf{83.4} & \textbf{0.98}
& \underline{\textbf{85.5}} & \underline{\textbf{0.96}} \\

CC-v2
& 72.1 & 1.24
& 78.5 & 1.02
& 72.0 & 1.28
& \underline{\textbf{70.6}} & \underline{\textbf{0.98}} \\

DC-v2
& 77.0 & 1.11
& 75.9 & 1.05
& \textbf{82.4} & \textbf{0.97}
& \underline{\textbf{84.8}} & \underline{\textbf{0.93}} \\

BC-v3
& 84.6 & 2.04
& 78.6 & 1.94
& 81.0 & 1.46
& \underline{\textbf{86.5}} & \underline{\textbf{0.98}} \\

CC-v3
& 70.7 & 1.18
& 73.2 & 1.26
& 66.5 & 1.13
& \underline{\textbf{64.2}} & \underline{\textbf{0.98}} \\

DC-v3
& 92.5 & 2.16
& 90.5 & 1.42
& \underline{\textbf{84.8}} & \underline{\textbf{0.99}}
& \textbf{82.0} & \textbf{0.96} \\

PB-v3
& 84.4 & 2.04
& 74.0 & 1.28
& \textbf{82.5} & \textbf{0.97}
& \underline{\textbf{83.4}} & \underline{\textbf{0.95}} \\

CB-v3
& 79.5 & 1.40
& 79.2 & 1.23
& 78.4 & 1.03
& \underline{\textbf{79.2}} & \underline{\textbf{0.98}} \\

PG-v3
& 88.6 & 1.58
& 86.8 & 1.15
& 88.5 & 1.38
& \underline{\textbf{86.8}} & \underline{\textbf{0.95}} \\

CG-v3
& 86.0 & 1.36
& 84.5 & 1.37
& \underline{\textbf{87.0}} & \underline{\textbf{0.99}}
& \textbf{85.2} & \textbf{0.97} \\

PP-v3
& 71.2 & 1.32
& 71.9 & 1.34
& 72.0 & 1.02
& \underline{\textbf{71.4}} & \underline{\textbf{0.96}} \\

CP-v3
& 84.3 & 1.78
& 81.2 & 1.01
& 86.8 & 1.08
& \underline{\textbf{82.3}} & \underline{\textbf{0.98}} \\

\midrule
Average
& 80.2 & 1.52
& 78.9 & 1.27
& 80.4 & 1.11
& \underline{\textbf{80.2}} & \underline{\textbf{0.97}} \\

\bottomrule
\end{tabular}
\captionsetup{
    width=\dimexpr\textwidth-4cm\relax,
    skip=4pt
}
\caption{Trajectory stitching ability analysis on multi-constraint tasks. Reward ($r$) is normalized to 0--100, and Cost-N ($c$) denotes the maximum normalized constraint violation. \textbf{Bold} indicates constraint satisfaction ($c \leq 1$), and \underline{\textbf{underlined bold}} indicates the highest reward among feasible methods.}
\label{tab:stitching_ability_scaled}
\end{table}

This subsection evaluates the trajectory stitching ability of different
sequence-modeling methods in multi-constraint offline safe reinforcement
learning. To construct a more challenging setting, we remove from the
offline datasets all complete trajectories that directly satisfy the
target safety-budget combinations. Consequently, the policy cannot
satisfy all constraints by simply imitating a complete jointly feasible
trajectory. Instead, it must learn complementary local behaviors from
different trajectories and compose them into a new safe decision
sequence during deployment. We compare CDT-MC, CORMA w/o PoE, CORDT,
and the full CORMA on Bullet-Safety-Gym and Safety-Gymnasium tasks.
The normalized reward measures task performance, while Cost-$N$
measures the maximum normalized violation among all constraints.

As shown in Table~\ref{tab:stitching_ability_scaled}, CDT-MC and CORMA w/o PoE achieve average normalized rewards of $80.2$ and $78.9$, respectively, but their average Cost-$N$ values reach $1.52$ and $1.27$, both clearly exceeding the safety threshold. CDT-MC satisfies all constraints only on BC-v2 and exhibits different degrees of safety violation on the remaining tasks. In particular, although CDT-MC achieves relatively high rewards on three-constraint tasks such as BC-v3, DC-v3, and PB-v3, its Cost-$N$ values reach $2.04$, $2.16$, and $2.04$, respectively. This indicates that its generated behaviors remain strongly biased toward reward maximization and cannot reliably compose local behaviors that satisfy different constraints. CORMA w/o PoE reduces the average Cost-$N$ to $1.27$, suggesting that the Mamba action decoder provides some benefit for long-horizon behavior modeling. Nevertheless, this variant fails to satisfy the safety threshold on all 12 tasks, indicating that conditioning Mamba only on multiple CTGs and a predefined RTG is insufficient to address multi-constraint behavior composition when jointly feasible trajectories are absent.

By introducing constraint-specific experts, a router, and PoE fusion, CORDT further reduces the average Cost-$N$ to $1.11$ and satisfies the safety threshold on five tasks: BC-v2, DC-v2, DC-v3, PB-v3, and CG-v3. Compared with CDT-MC and CORMA w/o PoE, these results indicate that explicitly modeling feasible RTG distributions under different constraints helps extract more effective high-level guidance from trajectories with complementary safety properties. For example, CORDT achieves reward--safety results of $84.8/0.99$ and $87.0/0.99$ on DC-v3 and CG-v3, respectively, demonstrating that it can effectively compose safe behaviors from different trajectories in some tasks. However, CORDT still violates the constraints on CC-v2, BC-v3, CC-v3, CB-v3, PG-v3, PP-v3, and CP-v3. This suggests that, with a Transformer-based action decoder, the fused RTG guidance cannot always be translated into temporally consistent safe actions over long horizons.

In contrast, CORMA achieves the most stable safety performance while maintaining competitive task rewards. Its average reward is $80.2$, which is comparable to $80.2$ for CDT-MC and $80.4$ for CORDT, whereas its average Cost-$N$ is only $0.97$. More importantly, CORMA is the only method that satisfies the safety threshold on all 12 tasks, with Cost-$N$ consistently ranging from $0.93$ to $0.98$. This shows that the safety advantage of CORMA is not limited to a few favorable environments, but remains stable across both two-constraint and three-constraint tasks. In particular, on BC-v3, CC-v3, PG-v3, PP-v3, and CP-v3, all competing methods violate the constraints, whereas CORMA still achieves feasible reward--safety results of $86.5/0.98$, $64.2/0.98$, $86.8/0.95$, $71.4/0.96$, and $82.3/0.98$, respectively.Several tasks also illustrate the adaptive reward--safety trade-off achieved by CORMA. On CC-v3, CDT-MC and CORMA w/o PoE obtain rewards of $70.7$ and $73.2$, respectively, but their corresponding Cost-$N$ values are $1.18$ and $1.26$. CORMA obtains a lower reward of $64.2$, but successfully controls Cost-$N$ at $0.98$. Similarly, on DC-v3 and CG-v3, CORDT achieves slightly higher feasible rewards than CORMA, whereas CORMA preserves a larger safety margin. 

These results demonstrate that CORMA possesses a certain degree of
trajectory stitching ability. The constraint-specific experts extract
feasible return information for individual constraints from different
trajectories, while the router-guided PoE mechanism coordinates these
complementary signals according to the current state and remaining
safety budgets. The Mamba action decoder then converts the fused
high-level guidance into temporally consistent safe action sequences
over long horizons. Compared with CDT-MC, CORMA w/o PoE, and CORDT,
CORMA does not achieve safety by substantially sacrificing the average
reward. Instead, it provides a more stable reward--safety trade-off,
supporting its ability to stitch trajectories from incomplete and
constraint-conflicting offline data.
\subsection{F.3 Analysis of the Zero-Shot Constraints Generalization }

This subsection further evaluates the zero-shot generalization ability
of different methods under unseen safety-budget combinations.
All unseen safety-budget combinations considered in this experiment
are sampled from the slight-shift region.
Specifically, we directly deploy the trained policies to a series of
v3 tasks without any additional fine-tuning. The normalized reward is
used to measure task performance, while Cost-$N$ denotes the maximum
normalized violation among all constraints, where Cost-$N\leq1$
indicates that all constraints are satisfied.

As shown in Table~\ref{tab:zero_shot_comparison}, CDT-MC achieves the highest average reward of $85.2$, but its average Cost-$N$ reaches $1.85$, and it fails to satisfy the safety threshold on all nine tasks. In particular, its Cost-$N$ values reach $2.15$, $2.14$, and $2.50$ on BC-v3, DC-v3, and CG-v3, respectively. These results indicate that, although CDT-MC can maintain high task returns under unseen budgets, its policy remains strongly biased toward reward maximization and cannot reliably coordinate multiple simultaneously changing safety constraints. Directly feeding multiple CTG conditions into a Transformer-based sequence model is therefore insufficient to guarantee safe generalization to unseen multi-dimensional budget combinations.

CORMA w/o PoE reduces the average Cost-$N$ from $1.85$ to $1.48$, demonstrating that the Mamba action decoder provides certain advantages for long-horizon conditional decision-making. Nevertheless, this variant satisfies the safety threshold only on PG-v3 and PP-v3, while constraint violations remain on the other seven tasks. In particular, its Cost-$N$ values reach $2.33$ and $1.82$ on DC-v3 and CG-v3, respectively. This suggests that, although Mamba improves long-range dependency modeling, it remains insufficient for handling conflicts among multiple heterogeneous constraints under unseen budget combinations without explicit constraint-specific RTG estimation and adaptive fusion.

CORDT further reduces the average Cost-$N$ to $1.20$ and achieves feasible performance on PB-v3, PG-v3, and CG-v3. Compared with CDT-MC and CORMA w/o PoE, these results demonstrate that constraint-specific experts, the router, and PoE fusion improve safety adaptation under unseen budgets. For example, CORDT achieves a reward--safety result of $89.7/0.99$ on CG-v3, satisfying all constraints while maintaining a high return. However, CORDT still violates the safety threshold on BC-v3, CC-v3, DC-v3, CB-v3, PP-v3, and CP-v3. This indicates that, even with constraint-aware high-level RTG guidance, the Transformer-based action decoder cannot consistently translate such guidance into safe long-horizon behavior across different tasks.

In contrast, CORMA keeps Cost-$N$ below $1.0$ on all nine tasks and is the only method that achieves constraint satisfaction across the entire evaluation suite. Its average Cost-$N$ is $0.97$, substantially lower than the values of $1.85$, $1.48$, and $1.20$ obtained by CDT-MC, CORMA w/o PoE, and CORDT, respectively. Meanwhile, CORMA achieves an average reward of $82.4$, which is only $2.8$ points lower than that of CDT-MC and is higher than those of CORMA w/o PoE and CORDT. These results show that the safety advantage of CORMA does not arise from overly conservative behavior. Instead, CORMA substantially improves the stability of constraint satisfaction while maintaining competitive task performance.

At the task level, CORMA achieves feasible reward--safety results of $80.2/0.98$, $72.0/0.97$, and $85.3/0.99$ on BC-v3, CC-v3, and DC-v3, respectively, whereas all competing methods violate the constraints on these three tasks. On CB-v3, PP-v3, and CP-v3, CORDT also fails to satisfy the safety threshold, while CORMA maintains Cost-$N$ values of $0.98$, $0.97$, and $0.94$, respectively. These results demonstrate that, even when multiple safety budgets simultaneously deviate from the training distribution, CORMA can still generate more reliable RTG guidance based on the current state and remaining safety budgets.

Several tasks further illustrate the adaptive reward--safety trade-off achieved by CORMA. On CC-v3, CDT-MC obtains a high reward of $86.7$, but its Cost-$N$ reaches $1.65$. In comparison, CORMA adjusts the reward to $72.0$ while reducing Cost-$N$ to $0.97$. On CG-v3, CORDT achieves a slightly higher feasible reward of $89.7$, whereas CORMA obtains a reward of $88.9$ with a lower Cost-$N$ of $0.95$, thereby preserving a larger safety margin. Similarly, both CORDT and CORMA satisfy the constraints on PB-v3, but CORMA trades a small amount of reward for a lower Cost-$N$. 

These experimental results verify the complementary effects of
constraint-specific RTG modeling and Mamba-based sequential
decision-making. The constraint-specific experts independently predict
feasible RTG distributions under different constraints, while the
router-guided PoE mechanism adaptively fuses these predictions based on
the current state and normalized remaining safety budgets. The Mamba
action decoder then converts the fused high-level guidance into stable
long-horizon actions. Compared with CDT-MC, CORMA w/o PoE, and CORDT,
CORMA satisfies all constraints on every evaluated task while
maintaining a competitive average reward, demonstrating more reliable
zero-shot generalization to unseen multi-dimensional safety-budget
combinations.

\begin{table}[t]
\centering
\renewcommand{\arraystretch}{1.2}
\setlength{\tabcolsep}{6pt}
\begin{tabular}{lcccccccc}
\toprule
\multirow{2}{*}{Task}
& \multicolumn{2}{c}{CDT-MC}
& \multicolumn{2}{c}{CORMA w/o PoE}
& \multicolumn{2}{c}{CORDT}
& \multicolumn{2}{c}{CORMA} \\
\cmidrule(lr){2-3}
\cmidrule(lr){4-5}
\cmidrule(lr){6-7}
\cmidrule(lr){8-9}
& Reward & Cost-$N$
& Reward & Cost-$N$
& Reward & Cost-$N$
& Reward & Cost-$N$ \\
\midrule

BC-v3
& 78.0 & 2.15
& 68.5 & 1.80
& 72.4 & 1.45
& \underline{\textbf{80.2}} & \underline{\textbf{0.98}} \\

CC-v3
& 86.7 & 1.65
& 77.2 & 1.41
& 75.5 & 1.28
& \underline{\textbf{72.0}} & \underline{\textbf{0.97}} \\

DC-v3
& 90.5 & 2.14
& 87.0 & 2.33
& 85.2 & 1.65
& \underline{\textbf{85.3}} & \underline{\textbf{0.99}} \\

PB-v3
& 78.2 & 1.99
& 76.4 & 1.54
& \underline{\textbf{75.8}} & \underline{\textbf{0.99}}
& \textbf{75.1} & \textbf{0.98} \\

CB-v3
& 85.6 & 1.59
& 84.8 & 1.39
& 84.2 & 1.08
& \underline{\textbf{84.0}} & \underline{\textbf{0.98}} \\

PG-v3
& 80.1 & 1.42
& \textbf{83.3} & \textbf{0.98}
& \textbf{81.5} & \textbf{0.97}
& \underline{\textbf{84.2}} & \underline{\textbf{0.98}} \\

CG-v3
& 92.4 & 2.50
& 90.8 & 1.82
& \underline{\textbf{89.7}} & \underline{\textbf{0.99}}
& \textbf{88.9} & \textbf{0.95} \\

PP-v3
& 85.9 & 1.46
& \textbf{81.4} & \textbf{0.99}
& 84.3 & 1.12
& \underline{\textbf{84.8}} & \underline{\textbf{0.97}} \\

CP-v3
& 89.5 & 1.78
& 89.6 & 1.06
& 88.2 & 1.27
& \underline{\textbf{87.5}} & \underline{\textbf{0.94}} \\

\midrule

Average
& 85.2 & 1.85
& 82.1 & 1.48
& 81.9 & 1.20
& \underline{\textbf{82.4}} & \underline{\textbf{0.97}} \\

\bottomrule
\end{tabular}
\captionsetup{
    width=\dimexpr\textwidth-4cm\relax,
    skip=4pt
}
\caption{Zero-shot generalization to unseen safety-budget combinations.
Reward is normalized to 0--100, and Cost-$N$ denotes the maximum
normalized constraint violation. \textbf{Bold} indicates constraint
satisfaction ($c\leq1$), and \underline{\textbf{underlined bold}}
indicates the highest reward among feasible methods.}
\label{tab:zero_shot_comparison}
\end{table}

\subsection{F.4 Effect of Trajectory Context Length on Stage-1 Action Decoder Training}
\begin{table}[t]
\centering
\renewcommand{\arraystretch}{1.2}
\setlength{\tabcolsep}{14pt}
\begin{tabular}{lcccc}
\toprule
\multirow{2}{*}{$K$} 
& \multicolumn{2}{c}{Transformer} 
& \multicolumn{2}{c}{Mamba} \\
\cmidrule(lr){2-3} \cmidrule(lr){4-5}
& $r\uparrow$ & $c\downarrow$
& $r\uparrow$ & $c\downarrow$ \\
\midrule
5  
& 89.8 & 2.44
& 88.5 & 2.21 \\

10 
& 86.1 & 2.12
& 82.4 & 1.96 \\

15 
& 85.9 & 1.96
& 83.6 & 1.68 \\

20 
& 80.7 & 1.61
& 77.6 & 1.42 \\

25 
& 79.4 & 1.58
& 75.0 & 1.24 \\

30 
& 75.8 & 1.15
& 74.5 & 1.10 \\

35 
& 79.4 & 1.37
& 72.9 & 1.03 \\

40 
& 80.3 & 1.58
& \textbf{73.3} & \textbf{0.99} \\

45 
& 83.8 & 1.76
& \textbf{74.7} & \textbf{0.96} \\

50 
& 83.2 & 1.93
& \underline{\textbf{75.1}} & \underline{\textbf{0.94}} \\

55 
& 84.9 & 2.10
& \textbf{74.9} & \textbf{0.95} \\

60 
& 85.6 & 2.24
& \textbf{74.6} & \textbf{0.97} \\
\bottomrule
\end{tabular}
\captionsetup{
    width=\dimexpr\textwidth-4cm\relax,
    skip=4pt
}
\caption{Effect of sequence length $K$ on Stage-1 action decoder
training. Results are averaged over three representative
three-constraint tasks: DC-v3, PB-v3, and CG-v3. Reward is
normalized to 0--100, and Cost-$N$ denotes the maximum normalized
constraint violation. \textbf{Bold} indicates constraint satisfaction
($c\leq1$), and \underline{\textbf{underlined bold}} indicates the
highest reward among feasible settings.}
\label{tab:k_length_stage1}
\end{table}

To investigate the effect of trajectory context length on Stage-1
action decoder training, we vary the input sequence length $K$ and
compare Transformer- and Mamba-based action decoders. The reported
results are averaged over three representative three-constraint
tasks, namely DC-v3, PB-v3, and CG-v3. Since $K$
determines the amount of historical state, action, RTG, and CTG
information available to the action decoder, this experiment examines
whether the two sequence-modeling backbones can effectively exploit
longer trajectory contexts under heterogeneous safety constraints.

As shown in Table~\ref{tab:k_length_stage1}, both action decoders
achieve relatively high rewards when the context length is small, but
these rewards are accompanied by severe constraint violations. At
$K=5$, the Transformer- and Mamba-based decoders obtain average
normalized rewards of $89.8$ and $88.5$, respectively, while their
average Cost-$N$ values reach $2.44$ and $2.21$. This result indicates
that short trajectory contexts do not contain sufficient information
for the action decoder to infer the long-term effects of current
actions on cumulative safety costs. Consequently, both models tend to
reproduce high-return behavior patterns without adequately controlling
multiple constraints.

Increasing the context length initially improves the safety
performance of both backbones, but their trends differ substantially.
For the Transformer-based decoder, the average Cost-$N$ decreases from
$2.44$ at $K=5$ to its lowest value of $1.15$ at $K=30$. However,
further increasing the context length causes safety performance to
deteriorate, with Cost-$N$ rising to $1.58$ at $K=40$ and $2.24$ at
$K=60$. Thus, although a moderately long context helps the Transformer
capture additional trajectory information, it does not provide stable
safety improvements as the input sequence becomes longer. Moreover,
the Transformer-based decoder fails to reduce the average Cost-$N$
below the safety threshold for any tested value of $K$.

In contrast, the Mamba-based decoder benefits more consistently from
longer trajectory contexts. Its average Cost-$N$ decreases from $2.21$
at $K=5$ to $1.10$ at $K=30$, and further falls below the normalized
safety threshold when $K\geq40$. At the same context length, Mamba also
achieves a lower average Cost-$N$ than the Transformer in every tested
setting. For example, at $K=40$, the Transformer-based decoder obtains
a Cost-$N$ of $1.58$, whereas the Mamba-based decoder reduces it to
$0.99$. These results suggest that the selective state-space mechanism
of Mamba can more effectively preserve safety-relevant information
over long horizons and exploit historical state--action--return--cost
dependencies for multi-constraint action generation.

The reward results further reveal a clear reward--safety trade-off.
As $K$ increases from $5$ to approximately $30$--$40$, the rewards of
both methods generally decrease while their safety performance
improves. This suggests that access to longer safety-related histories
encourages the action decoders to generate more cautious behavior.
Although the Transformer-based decoder obtains higher rewards than
Mamba for most context lengths, these results are consistently
associated with Cost-$N>1$ and therefore do not constitute feasible
reward--safety trade-offs. By comparison, the Mamba-based decoder
maintains an average normalized reward between approximately $73$ and
$75$ while keeping its average Cost-$N$ below $1.0$ for
$K\in\{40,45,50,55,60\}$.

Among all tested context lengths, $K=50$ provides the most favorable
average reward--safety trade-off for the Mamba-based action decoder,
achieving a normalized reward of $75.1$ and a Cost-$N$ of $0.94$.
Increasing $K$ further to $55$ or $60$ does not yield additional
benefits: the average reward slightly decreases to $74.9$ and $74.6$,
while Cost-$N$ increases to $0.95$ and $0.97$, respectively. This
suggests that excessively long contexts may introduce redundant or
less relevant historical information without improving action
generation. We therefore set $K=50$ in the subsequent experiments to
balance long-horizon dependency modeling, constraint satisfaction, and
task performance.

\subsection{F.5 How Critical Is the Three-Stage Training Strategy?}
\begin{table}[t]
\centering
\small
\renewcommand{\arraystretch}{1.1}
\setlength{\tabcolsep}{4.2pt}
\begin{tabular}{l *{5}{cc}}
\toprule
\multirow{2}{*}{\textbf{Method}}
& \multicolumn{2}{c}{\textbf{DC-v3}}
& \multicolumn{2}{c}{\textbf{PB-v3}}
& \multicolumn{2}{c}{\textbf{CG-v3}}
& \multicolumn{2}{c}{\textbf{CP-v3}}
& \multicolumn{2}{c}{\textbf{Average}} \\
\cmidrule(lr){2-3}
\cmidrule(lr){4-5}
\cmidrule(lr){6-7}
\cmidrule(lr){8-9}
\cmidrule(lr){10-11}
& $r\uparrow$ & $c\downarrow$
& $r\uparrow$ & $c\downarrow$
& $r\uparrow$ & $c\downarrow$
& $r\uparrow$ & $c\downarrow$
& $r\uparrow$ & $c\downarrow$ \\
\midrule

CORMA w/o 3-stage Training
& 91.3 & 2.86
& 92.1 & 2.24
& 88.5 & 2.65
& 89.7 & 2.18
& 90.4 & 2.48 \\

CORMA w/ Mean-Regression Router
& 84.7 & 1.31
& 88.6 & 1.18
& 85.1 & 1.35
& 85.9 & 1.20
& 86.1 & 1.26 \\

CORMA w/o Balanced Router Sampling
& 85.8 & 1.17
& 88.7 & 1.09
& 89.6 & 1.21
& 86.5 & 1.11
& 87.7 & 1.15 \\

CORMA w/o Expert Freezing
& 83.1 & 1.45
& 87.1 & 1.28
& 83.2 & 1.48
& 82.5 & 1.30
& 84.0 & 1.38 \\
\midrule

\textbf{CORMA (Ours)}
& \textbf{88.0} & \textbf{0.95}
& \textbf{90.2} & \textbf{0.94}
& \textbf{91.1} & \textbf{0.90}
& \textbf{88.3} & \textbf{0.95}
& \textbf{89.4} & \textbf{0.94} \\

\bottomrule
\end{tabular}
\captionsetup{
    width=\dimexpr\textwidth-4cm\relax,
    skip=4pt
}
\caption{Ablation study on the three-stage optimization strategy
of CORMA across four multi-constraint tasks. Reward ($r$) is
normalized to 0--100, and Cost-$N$ ($c$) denotes the maximum
normalized constraint violation.}
\label{tab:training_ablation}
\end{table}
To evaluate the effectiveness and generality of the training
mechanism in CORMA, we conduct an ablation study on four challenging
three-constraint tasks: DC-v3, PB-v3, CG-v3, and CP-v3. We compare
the full CORMA with four variants:
(1) w/o 3-stage Training, which jointly optimizes the action decoder,
constraint-specific experts, and router;
(2) w/ Mean-Regression Router, which replaces the PoE
negative log-likelihood in Stage III with an MSE loss on the fused
RTG mean;
(3) w/o Balanced Router Sampling, which samples the
constraint-specific subsets according to their original data
proportions rather than with equal probability; and
(4) w/o Expert Freezing, which updates the expert parameters together
with the router during Stage III.
  
As shown in Table~\ref{tab:training_ablation}, the full CORMA achieves the most stable reward--safety trade-off across all four tasks, obtaining an average normalized reward of $89.4$ while maintaining an average Cost-$N$ of $0.94$. More importantly, it is the only variant that satisfies all constraints on every evaluated task.Jointly optimizing the action decoder, experts, and router produces the highest average reward of $90.4$, but increases the average Cost-$N$ to $2.48$. This substantial safety degradation indicates that simultaneous optimization exposes the router to continuously changing expert distributions and introduces interference among the objectives of the three modules.Replacing the PoE likelihood with mean regression reduces the average Cost-$N$ to $1.26$, but still fails to satisfy the safety threshold on all four tasks. This result suggests that aligning only the fused RTG mean is insufficient, because the router cannot exploit the predictive variances of the constraint-specific expert distributions.Removing balanced router sampling results in an average Cost-$N$ of $1.15$. Although this variant performs better than joint training and mean regression, it remains unsafe across all evaluated tasks. This confirms that sampling the constraint-specific subsets according to their natural frequencies may cause constraints with larger feasible subsets to dominate router optimization.Unfreezing the experts during Stage III further degrades both reward and safety, yielding an average reward of $84.2$ and an average Cost-$N$ of $1.38$. Updating the experts together with the router causes the constraint-specific RTG distributions learned in Stage II to drift, thereby creating non-stationary targets for PoE fusion.

Overall, these results demonstrate that sequential module
specialization, distributional PoE supervision, balanced router
sampling, and expert freezing are jointly important for achieving
stable multi-constraint RTG estimation and constraint satisfaction.
\end{adjustwidth}

\end{document}